\documentclass[10pt,journal,compsoc]{IEEEtran}

\usepackage{amsmath}
\usepackage{amssymb}
\usepackage{mathtools}
\usepackage{amsthm}

\usepackage{hyperref}
\usepackage{microtype}
\usepackage{graphicx}
\usepackage{subfigure}
\usepackage{booktabs} % for professional tables
\usepackage{multirow}
\usepackage{makecell}
\usepackage{hyperref}
\usepackage{xcolor}
\usepackage{colortbl}
\usepackage{paralist}
\usepackage[textsize=tiny]{todonotes}
\usepackage{tabularx} 
\usepackage{arydshln}
\usepackage{algorithm}
\usepackage[noend]{algorithmic}
\usepackage{setspace}

\ifCLASSOPTIONcompsoc
  \usepackage[nocompress]{cite}
\else
  \usepackage{cite}
\fi

\ifCLASSINFOpdf

\else

\fi

\begin{document}

\title{Just Noticeable Difference for Large Multimodal Models}

\author{Zijian~Chen, Yuan Tian, Yuze Sun, Wei Sun,~\IEEEmembership{Member,~IEEE,} Zicheng Zhang, \\Weisi~Lin,~\IEEEmembership{Fellow,~IEEE,} Guangtao~Zhai,~\IEEEmembership{Fellow,~IEEE,}
        and~Wenjun~Zhang,~\IEEEmembership{Fellow,~IEEE,}
        % <-this % stops a space
\IEEEcompsocitemizethanks{\IEEEcompsocthanksitem Zijian Chen, Yuan Tian, Zicheng Zhang, and Guangtao Zhai are with the Institute of Image Communication and Information Processing, Shanghai Jiao Tong University, Shanghai 200240, China, and also with the Shanghai AI Laboratory, Shanghai 200232, China (e-mail: \{zijian.chen, ee\_tianyuan, zzc1998, zhaiguangtao\}@sjtu.edu.cn). (\textit{Corresponding author: Guangtao Zhai.})
}
\IEEEcompsocitemizethanks{\IEEEcompsocthanksitem Yuze Sun and Wenjun Zhang are with the Institute of Image Communication and Information Processing, Shanghai Jiao Tong University, Shanghai 200240, China (e-mail: \{StevenSun, zhangwenjun\}@sjtu.edu.cn).
}
\IEEEcompsocitemizethanks{\IEEEcompsocthanksitem Wei Sun is with the School of Communication and Electronic Engineering, East China Normal University, Shanghai 200241, China (e-mail: wsun@cee.ecnu.edu.cn).
}
\IEEEcompsocitemizethanks{\IEEEcompsocthanksitem \fontdimen2\font=0.5ex Weisi Lin is with the School of Computer Science and Engineering, Nanyang Technological University, Singapore 639798 (e-mail: wslin@ntu.edu.sg).
}
}

%\fontdimen2\font=0.5ex
% To what level of granularity can large multimodal models (LMMs) perceive details? Existing LMMs have made remarkable breakthroughs in many vision tasks such as grounding, image quality assessment, and cross-scenario reasoning. However, the minimal perceptible change of LMMs remains unclear. demonstrate that the LMM has the JND, termed as the LMM-JND.

%Our research reveals that there exist significant visual blind spots in current LMMs. To systemically quantify this characteristic, we propose a novel concept, {\bf LMM-JND}, namely the Just Noticeable Difference together with its determination pipeline for LMMs.

%Specifically, 

%we also construct the visual perception alignment JND dataset (VPA-JND), containing 21.5k reference images with over 489k stimuli across 12 distortion types, to facilitate this research.

% With these pairs, we identify the number of LMM-blind pairs at different levels of signal variation that LMMs respond as equivalent despite their visual differences.

\markboth{Preprint}%
{Shell \MakeLowercase{\textit{et al.}}: Bare Demo of IEEEtran.cls for Computer Society Journals}

\IEEEtitleabstractindextext{%
\begin{abstract}
Just noticeable difference (JND), the minimum change that the human visual system (HVS) can perceive, has been studied for decades. Although recent work has extended this line of research into machine vision, there has been a scarcity of studies systematically exploring its perceptual boundaries across multiple tasks and stimulus types, particularly in the current era of rapidly advancing large multimodal models (LMMs), where studying the multifaceted capabilities of models has become a mainstream focus.
Moreover, the perceptual defects of LMMs are not investigated thoroughly, resulting in potential security issues and suboptimal response efficiency.
In this paper, we take an initial attempt and demonstrate that there exist significant visual blind spots in current LMMs. To systemically quantify this characteristic, we propose a new concept, {\bf LMM-JND}, together with its determination pipeline.
Targeting uncovering the behavior commonalities in HVS-aligned visual perception tasks, we delve into several LMM families and construct a large-scale dataset, named VPA-JND, which contains 21.5k reference images with over 489k stimuli across 12 distortion types, to facilitate LMM-JND studies.
VPA-JND exposes areas where state-of-the-art LMMs, including GPT-4o and the InternVL2.5 series, struggle with basic comparison queries and fall significantly short of human-level visual performance.
We further explore the effects of vision and language backbones and find a notable correlation between their design philosophy that may instruct the future refinement of LMMs for their visual acuity.
Together, our research underscores the significance of LMM-JND as a unique perspective for studying LMMs, and predictable LMM-JND is crucial for security concerns. This work will be available at \url{https://github.com/zijianchen98/LMM-JND}.
\end{abstract}

}

% make the title area
\maketitle

\IEEEdisplaynontitleabstractindextext

\IEEEpeerreviewmaketitle

\IEEEraisesectionheading{
\section{Introduction}
\label{Intro}}

\IEEEPARstart{R}{ecent} years have witnessed unprecedented developments in large multimodal models (LMMs) \cite{GPT-4o, li2024llava, gemini2.5 ,bai2025qwen25vl,claude}, driven by advancements in architectural innovation, cross-modal alignment techniques, and different training paradigms. Their powerful visual and textual data integration capabilities showcase excellent performance in all-round tasks such as image quality assessment (IQA) \cite{wu2024qbench,chen2024gaia}, visual question answering (VQA) \cite{chen2024obi,fu2023mme}, and causal reasoning \cite{yue2024mmmu}.  
In particular, the recent InternVL3 \cite{zhu2025internvl3} series have pushed performance closer to or even exceeded proprietary models.

\begin{figure}[t]
\begin{center}
\centerline{\includegraphics[width=1\columnwidth]{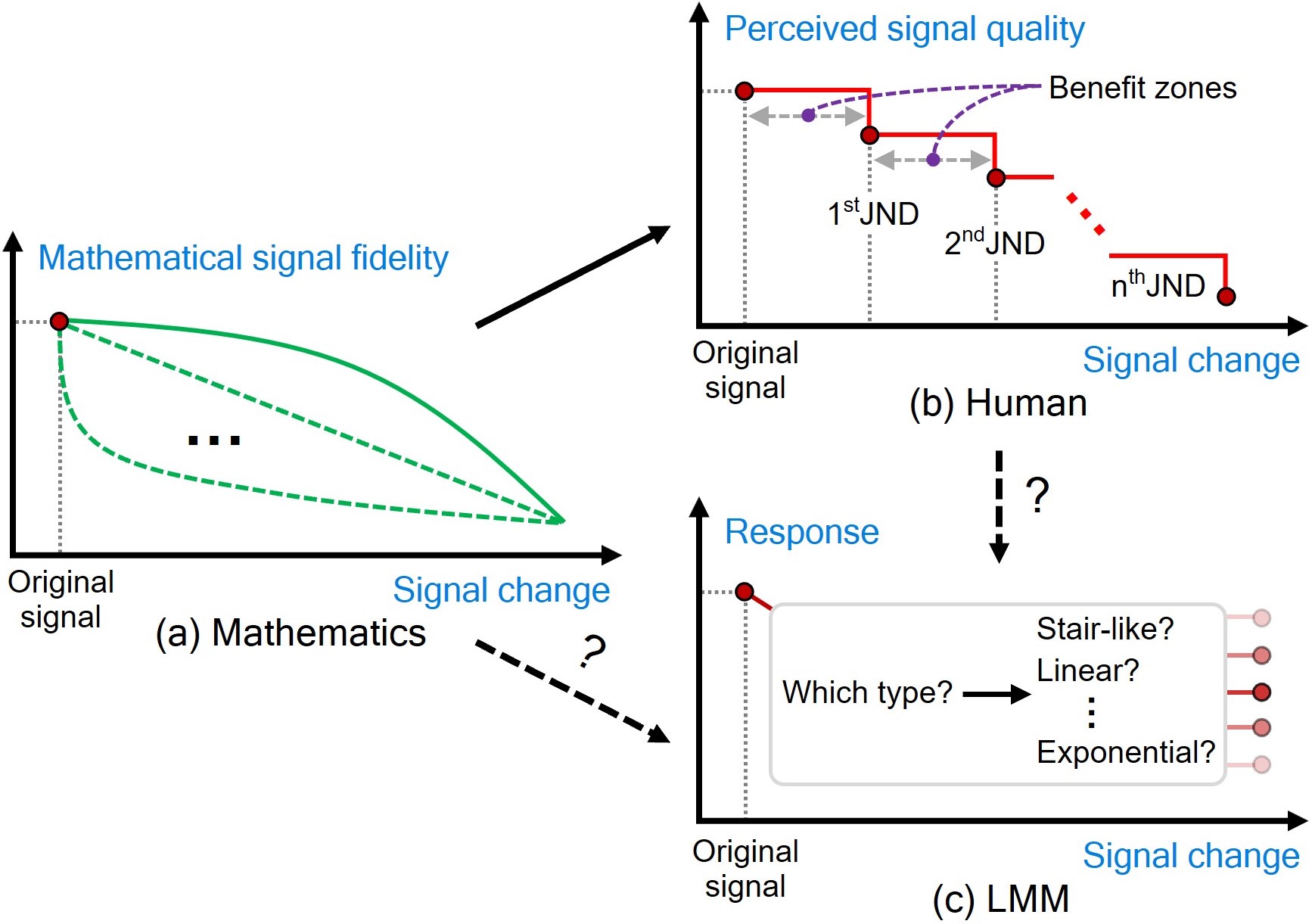}}

\caption{Illustration of different JND modelings. Mathematically, the fidelity of a signal varies with its content according to a specific function ({\it e.g.}, linear, exponential, or logarithmic). While humans perceive signal quality with benefit zones due to the physiological boundaries. {\it Do LMMs respond likewise?}}
\label{intro}
\end{center}

\end{figure}

\begin{figure*}[t]
\begin{center}
\includegraphics[width=1\linewidth]{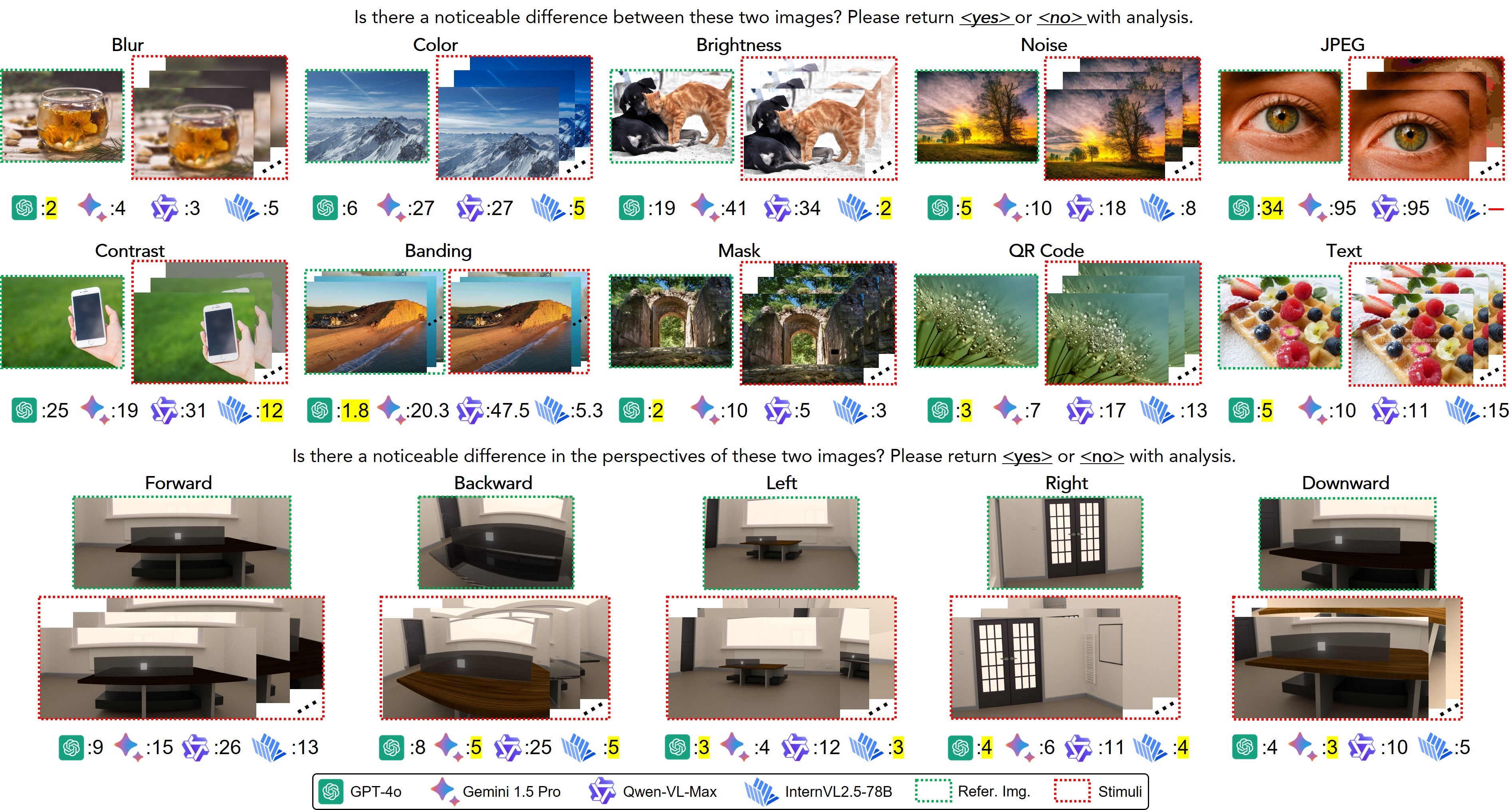}
\caption{Examples of the minimal distortion level that can be perceived ({\it i.e.} $1$\textsuperscript{st} JND point). The smallest perceptible levels are shaded in yellow. The red short dash `\textcolor{red}{\textendash}' indicates that the minimum perceptible range exceeds the maximum distortion level. We can observe that both leading proprietary LMMs (GPT-4o, Gemini 1.5 Pro, and Qwen-VL-Max) and open-source LMM (InternVL2.5-78B) exhibit varying degrees of visual granularity on different stimuli (See a more comprehensive comparison in Tab. \ref{exp1}).}
\label{example}
\end{center}
\end{figure*}

Beyond these advancements, few studies have investigated the perception limits of LMMs, which not only include the breadth of their capabilities but also their granularity. As the ultimate receiver and appreciator of the increasingly larger number of visual contents change from the human visual system (HVS) to LMMs, the concerns regarding their perceptual capabilities and safety are gaining prominence. Fig. \ref{example} shows that many LMMs struggle to detect and respond differently to changes that humans can easily perceive. 
Thus, a question naturally raises: {\it what is the minimal magnitude of changes that LMMs can perceive?}
Such a characteristic is also known as the just noticeable difference (JND) that has been widely explored with the unique psychological and physiological mechanisms of HVS during the past decades \cite{shen2003foundations,lin2021progress}.
Leveraging the benefit zones in HVS (Fig. \ref{intro}), HVS-JND plays an important role in multimedia data production, representation, and transmission \cite{wu2013perceptual,chen2014jnd,liu2019deep,yuan2019visual}.
More recently, researchers have extended this concept to machine vision system (MVS) \cite{zhang2021just,jin2022just,zhang2023learning} and verified the existence of JND on the image classification and object detection tasks, resulting in feasible optimization to machine-oriented visual coding and promoting the implementation of machine vision applications in various scenarios.

However, the characteristics obtained from HVS-oriented JND and MVS-oriented JND may not be directly transferred to LMMs due to the following facts. 1) Structural differences of the signal processing pathway: Leaving aside the biological and physiological structures of the HVS, traditional single-modal vision models, such as convolutional neural networks (CNN) and Transformers, differ significantly from current LMMs.
The latter augments a foundational visual encoder with a cross-modal alignment module and a decoder-only language generation backbone, which normally have several orders of magnitude of large parameter quantities.
2) Task-specificity: Previous studies \cite{jin2022just,zhang2023learning} defined the MVS-JND mainly based on the outcome of simple visual tasks, while the application scope of large models is obviously broader and exhibiting prompt-dependent characteristic.
3) Lack of relevant research: To the best of our knowledge, the existence and characteristics of JND in LMMs remain under-confirmed. Although some works \cite{tong2024eyes,cui2024robustness,yang2025visionzip} have explored the visual shortcomings of LMMs from the perspectives of perception ability, visual token redundancy, and robustness, their specific magnitude and form have not been systematically investigated.
If LMMs have JND, there may exist several potential impacts to take into account. 
First, finding the LMM-JND points will greatly benefit the processing efficiency of visual content.
JND points act as a {\it benefit zone} that allows the system to ignore insignificant or irrelevant variations in the input, thus helping compress information, refine the model structure, enhance routine user interactions for online platforms, improve the constantly expanding benchmarks, {\it etc}.
Meanwhile, such perceptual redundancy features could also explain security concerns, {\it e.g.}, data tampering and attacking. Therefore, the presence of LMM-JND precisely indicates the robustness of the LMM in certain scenarios.

\begin{table*}[t]
\renewcommand\arraystretch{1.05}
 \centering
\caption{Comparison of Existing JND-based Datasets with VPA-JND. {\it Combination\textsuperscript{N}} Denotes that the Dataset Contains $N$ Different Distortion Types. \textsuperscript{*}Machine Denotes that Only Neural Network Models for Image Classification or Object Detection Tasks are Tested. \textsuperscript{$\dag$}We Report the Number of Unique Reference Images. All Datasets are Listed in Chronological Order of Publication}
\label{dataset_comparison}
\resizebox{1\linewidth}{!}{\begin{tabular}{lccccccccc}
\toprule
Dataset &Year &Stimuli Type & \# Ref. &\# Dist. Level &\# Total Stimuli &Resolution &Distortion Type&Test Object \\
\midrule
 MCL-JCI \cite{jin2016statistical}&2016&Image&50&100&5,050&1920$\times$1080&JPEG& Human\\
  MCL-JCV \cite{wang2016mcl}&2016&Video&30&51&1,560&1920$\times$1080& H.264/AVC&Human\\
   JND-HEVC \cite{huang2017measure}&2017&Video&40&51&2,080&1920$\times$1080&H.265/HEVC&Human\\
   VideoSet \cite{wang2017videoset}&2017&Video&220&51&45,760&1920$\times$1080&H.264/AVC&Human\\
    JND-Pano \cite{liu2018jnd}&2018&Image&40&100&4,040&5000$\times$2500&JPEG&Human\\
     SIAT-JSSI \cite{fan2019picture}&2019&Image&12&\{300, 51\}&4,040&1920$\times$1080& \makecell[c]{JPEG2000,\\ H.265/HEVC}&Human\\
       VVC \cite{shen2020just}&2020&Image&202&39&7,878&1920$\times$1080&H.266/VVC& Human\\
        KonJND-1k \cite{lin2022large}&2022&Image&1,008&\{100, 51\}&77,112&640$\times$480&JPEG, BPG&Human \\
        DMV-JND \cite{jin2022just}&2022&Image&\textendash&\textendash&\textendash&32$\times$32&DMV-JND distortion&\textsuperscript{*}Machine\\
            R3VIVAL \cite{klein2023r3vival}&2023&Video, audio&34&8&272&4096$\times$2160&Absorption&Human\\
       JPEG AIC-3 \cite{testolina2023jpeg}&2023&Image&10&\textendash&500&945$\times$880&{\it Combination\textsuperscript{5}}&Human \\
       CTF-JND \cite{liu2023first}&2023&Image&106&\textendash&1,642&\textendash&{\it Combination\textsuperscript{25}}&Human\\
       SMR \cite{zhang2024perceptual} &2024&Image&740,766&36&27,408,342&\textendash&H.265/HEVC&\textsuperscript{*}Machine\\
\midrule
 \rowcolor{cyan!10} 
 {\bf VPA-JND} (Ours)&2025&Image&\textsuperscript{$\dag$}21,598&\{50, 100\}&489,065&224\textsuperscript{2}-1080p&{\it Combination\textsuperscript{14}}&LMM\\
\bottomrule
\end{tabular}}
\end{table*}

In this paper, we make the initial attempt to study JND for LMMs to tackle these challenges and to pave the way for future research upon LMM-JND-guided tasks. Our core contributions are summarized as follows.
\begin{itemize}
\item We demonstrate an uninvestigated area in the vision and language backbones of LMMs through two pilot studies. The first one reveals that some vision backbones show close feature representations under different distortion levels. The second one indicates a dumb commonality shared across existing LMMs, which manifests as disproportionate response sensitivity to image distortions.
\item We propose a novel concept, LMM-JND, to quantify the perceptual redundancy characteristic for LMMs and a well-designed pipeline for its determination. LMM-JND is defined as the minimal amount of stimulus that must be changed for an input difference to be perceived by an LMM. To the best of our knowledge, our work is the first to explore JND in LMMs.
\item We study LMM-JND across representative LMM families focusing on three types of perceptual stimuli including low-level distortions, content-injection, and 3D spatial variations. We build a large-scale LMM-oriented just noticeable difference image dataset ({\bf VPA-JND}) containing over 489k stimuli with more than 21k reference images. This dataset fosters further JND studies on LMMs.
\item We conduct extensive experiments to uncover the JND characteristic at the model level and the stimulus level. We find LMMs exhibit different JND characteristics on specific stimuli while still falling far short of human-level capabilities in detail perception.
We also notice that the significant commonalities of JND observed in the HVS are attenuated in large models of varying scales and architectures, endowing them with more pronounced individual characteristics and stronger task dependency.
\item We delve into the architecture of LMMs and discover a special correlation between the scale ratio of their language and vision backbones and the models' perceptual thresholds, providing possible optimization directions for future LMMs.
\end{itemize}

\section{Related Works}

\subsection{Subjective JND Studies and Datasets} 
Numerous JND datasets focusing on different compression schemes and multimedia contents have been proposed to evaluate human subjective perception in the past decade \cite{wu2019survey,lin2021progress}. Early representative JND datasets, such as MCL-JCI \cite{jin2016statistical}, MCL-JCV \cite{wang2016mcl}, VideoSet \cite{wang2017videoset}, JND-HEVC \cite{huang2017measure}, and VVC \cite{shen2020just}, investigate the JPEG distortion as well as H.264/AVC, H.265/HEVC, and H.266/VVC video coding distortions in the context of image and video compression, respectively. Subsequently, KonJND-1k \cite{lin2022large} further expanded the scale of the JND dataset to meet the prevalence of data-driven JND estimation models.
As distortion diversifies and visual content becomes more complex, researchers have shifted their focus to developing JND datasets tailored to emerging content types, as represented by JND-Pano \cite{liu2018jnd}, and SIAT-JSSI \cite{fan2019picture}, which were collected for panoramic and stereoscopic images, respectively. Following this, a generalized JND dataset \cite{liu2023first} comprising 25 distortion types was constructed through a coarse-to-fine JND selection. In addition, Sheng {\it et al.} \cite{sheng2024audio} have conducted an audio-video collaborative JND research, which involves cross-sensory JND point collection.
Nevertheless, these studies are predominantly HVS-oriented and used for refining multimedia systems ({\it e.g.}, transmission and display). 
On the other hand, a few researchers \cite{zhang2021just, jin2022just, zhang2024perceptual} have demonstrated the existence of JND for machines and constructed corresponding datasets ({\it e.g.} DMV-JND \cite{jin2022just} and SMR \cite{zhang2024perceptual}) to explore their distortion tolerance properties for machine-oriented visual coding. 
In this work, we further transition the idea of JND to the prevailing LMMs, and propose the VPA-JND dataset to evaluate the perception redundancy characteristics of LMM.
A detailed comparison between existing JND datasets is in Tab. \ref{dataset_comparison}. 

\begin{table*}
    \centering
    \renewcommand\arraystretch{1.2}
    \renewcommand\tabcolsep{4pt}
        \caption{A Brief View of 16 Open-Source LMMs Evaluated in this Work in Chronological Order. `\# Token' Denotes the Number of Visual Tokens Fed into the LLM. `GPU Mem.' Represents the VRAM Usage when Processing an Image Pair with $640\times 480$ Resolution}
    \resizebox{1\linewidth}{!}{\begin{tabular}{|c|l|cc|cc|cc|c|}
    \hline
        \multirow{2}{*}{Index}&\multirow{2}{150pt}{$^\textit{Date of Release}$Model Names} & \multicolumn{2}{c|}{Vision Architectures (V)}  & \multicolumn{2}{c|}{V$\to$L} &\multicolumn{2}{c|}{Language Architectures  (L)}& \multirow{2}{*}{GPU Mem.}\\ \cdashline{3-8}
       & & Backbone & \# Size &  Alignment& \# Token & Backbone & Type&   \\ 
        \hline
      A &  $^\textit{25.01}$SmolVLM-256M \cite{SmolVLM2025}&ViT-93M&512&\textendash&64&SmolLM2-135M&{\it decoder-only}&2.02 GB\\
     B &   $^\textit{25.01}$Qwen2.5-VL-72B \cite{bai2025qwen25vl}&QwenViT-675M&224&MLP projector&\textendash& Qwen2.5-72B&{\it decoder-only}&138.99 GB\\
     C &   $^\textit{25.01}$Qwen2.5-VL-7B \cite{bai2025qwen25vl}&QwenViT-675M&224&MLP projector&\textendash&Qwen2.5-7B&{\it decoder-only}&16.72 GB\\
     D  &  $^\textit{25.01}$Qwen2.5-VL-3B \cite{bai2025qwen25vl}&QwenViT-675M&224&MLP projector&\textendash&Qwen2.5-3B&{\it decoder-only}&9.47 GB\\
    E  &    $^\textit{24.12}$DeepSeek-VL2 (27.5B) \cite{wu2024deepseek}&SigLIP-400M&384&MLP projector&421& DeepSeekMoE-27B&{\it decoder-only}&83.88 GB\\
    F  &   $^\textit{24.12}$DeepSeek-VL2-Small (16.1B) \cite{wu2024deepseek}&SigLIP-400M&384&MLP projector&421&DeepSeekMoE-16B&{\it decoder-only}&52.35 GB\\
    G &    $^\textit{24.12}$DeepSeek-VL2-Tiny (3.4B) \cite{wu2024deepseek}&SigLIP-400M&384&MLP projector&421&DeepSeekMoE-3B&{\it decoder-only}&22.79 GB\\
    H &   $^\textit{24.12}$InternVL2.5-78B \cite{chen2024expanding}&InternViT-6B-v2.5&448&MLP+Cross Attn.&256&Qwen2.5-72B&{\it decoder-only}&154.34 GB\\
     I &   $^\textit{24.12}$InternVL2.5-38B \cite{chen2024expanding}&InternViT-6B-v2.5&448&MLP+Cross Attn.&256&Qwen2.5-32B&{\it decoder-only}&81.68 GB\\
     J &   $^\textit{24.12}$InternVL2.5-8B \cite{chen2024expanding}&InternViT-300M-v2.5&448&MLP+Cross Attn. &256&InternLM2.5-7B&{\it decoder-only}&24.92 GB\\
     K &  $^\textit{24.09}$Qwen2-VL-72B \cite{wang2024qwen2}&QwenViT-675M&224&MLP projector&66& Qwen2-72B&{\it decoder-only}&138.95 GB\\
    L &    $^\textit{24.09}$Qwen2-VL-7B \cite{wang2024qwen2}&QwenViT-675M&224&MLP projector&66&Qwen2-7B&{\it decoder-only}&16.77 GB\\
     M &   $^\textit{24.09}$Qwen2-VL-2B \cite{wang2024qwen2}&QwenViT-675M&224&MLP projector&66&Qwen2-1.5B&{\it decoder-only}&5.42 GB\\
     N &   $^\textit{24.08}$LLaVA-OneVision-72B \cite{li2024llava}&SigLIP-400M&384&MLP projector&729&Qwen2-72B&{\it decoder-only}&142.94 GB\\
    O &   $^\textit{24.08}$LLaVA-OneVision-7B \cite{li2024llava}&SigLIP-400M&384& MLP projector&729&Qwen2-7B& {\it decoder-only}&17.47 GB\\
    P&     $^\textit{24.08}$LLaVA-OneVision-0.5B \cite{li2024llava}&SigLIP-400M&384& MLP projector&729&Qwen2-0.5B &{\it decoder-only}&3.11 GB\\
        \hline
    \end{tabular}}
    \label{tab:arch}
\end{table*}

\subsection{Visual JND Models and Applications}
There have been substantial studies in the HVS-JND during the past decades \cite{chou1995perceptually,yuan2019visual}, which can be broadly divided into two categories: pixel- and subband-domain. The former directly computes the HVS-JND threshold of each pixel by considering background luminance and various masking effects drawn from psychophysical experiments \cite{brown1989high,pelli2013measuring}. The latter first transfers the pixel-domain image into the subband-domain one using transformations, {\it e.g.}, discrete cosine transform (DCT) \cite{bae2013novel}, discrete wavelet transform (DWT) \cite{bradley1999wavelet}, or Karhunen-Loève transform (KLT) \cite{jiang2022toward}, and then estimates the HVS-JND via contrast sensitivity function (CSF) \cite{wei2009spatio}, luminance adaptation \cite{ahumada1992luminance}, or textural masking \cite{liu2010just}.
Recently, some learning-based HVS-JND models have been proposed \cite{liu2019deep,zhang2021deep,mao2023transfer} to fill the gap caused by the limited understanding of the HVS. 
The HVS-JND model is typically employed to characterize the visual redundancy present in images and videos, making it widely applicable in perceptual signal processing domains such as image and video compression \cite{wang2020hierarchical,cao2024sg}, digital watermarking \cite{wan2020pattern}, and image quality assessment (IQA) \cite{flynn2013image, ak2022just}.
The effectiveness and benefits of HVS-JND techniques rely on the assumption that the HVS serves as the ultimate receiver.
Nowadays, an ever-growing volume of multimedia data is analyzed by LMMs.
To enhance the efficiency, robustness, and security of the LMM while expanding its applications, it is crucial to explore the redundancy of visual contents.
Therefore, the JND for LMMs should be studied and defined.

\subsection{Large Multimodal Models} 
LMMs have achieved profound impact and revolution on the entire AI community and beyond \cite{yin2024survey,zhang2024mm}, where a bunch of proprietary and open-source LMMs is developed to tackle multimodal inputs ({\it e.g.}, text, image, audio, point cloud, and code) and tasks ({\it e.g.}, perception, understanding, reasoning, and planning) \cite{GPT-4o, reid2024gemini, llama32vision, chen2024expanding, wang2024qwen2}. Building on top of these capabilities, researchers further apply LMMs to advance specific research areas.
For example, Q-Bench \cite{wu2024qbench} proposes a binary softmax strategy, enabling LMMs to predict quantifiable quality scores for visual content. DriveGPT4 \cite{xu2024drivegpt4} interprets video inputs to generate driving-related textual responses with pertinent reasoning.
OBI-Bench \cite{chen2024obi} demonstrates the potential of LMMs in deciphering ancient scripts.
Med-MLLM \cite{liu2023medical} leverages medical data across visual and textual modalities ({\it e.g.}, chest X-ray and free-text clinical note) for epidemiological diagnosis.
Without exception, all these applications involve visual comparisons, which deeply rely on the perception granularity of LMMs.
However, the absence of a comprehensive investigation into this attribute across LMMs with different scales or architectures limits their reliability, which also poses safety hazards for LMMs themselves.
Our work takes the inherent correlation of the above concerns and commences by introducing the JND of LMMs.

\begin{table*}[t]
    \centering
    \renewcommand\arraystretch{1.05}
    \caption{Comparison of the Response Sensitivity Differences of LMMs in Different Distortion Intervals. In the `HVS-JND' Column, We Report the Absolute/Relative Perceptible Distortion Level. $\Delta$ Denotes the Range of Values. The Results of `\# Response Variation Label' are Averaged from 50 Reference Images with 2,500 Stimuli Three Times to Mitigate Model Hallucinations}
    \resizebox{1\linewidth}{!}{\begin{tabular}{c|c|c|c|ccccccccccccccccc}
    \hline                 
        \multirow{2}{*}{Dist. level}&\multirow{2}{*}{$\Delta$PSNR (dB)}&\multirow{2}{*}{$\Delta$SSIM}& \multirow{2}{*}{HVS-JND}&\multicolumn{16}{c}{\# Response
Variation Label}  \\ \cdashline{5-20}
        &&&&A&B&C&D&E&F&G&H&I&J&K&L&M&N&O&P\\
        \hline
        0$\to$5&[32.07, 33.45]&[0.9621, 0.9671]&$>$5/$>$5&0&0&0&.752&0&0&0&.104&.304&.268&0&0&0&.012&.020&0\\
         5$\to$10&[34.71, 48.44]&[0.9937, 0.9989]&8.65/3.65&0&0&0&.824&0&0&0&.136&.268&.308&0&0&0&.008&.004&0\\
         10$\to$20&[27.61, 47.67]&[0.9774, 0.9984]&12.33/2.33&0&.014&0&.916&0&0&0&.174&.310&.292&0&0&0&.042&.008&0\\
          20$\to$30&[29.36, 46.23]&[0.9706, 0.9978]&22.87/2.87&0&.196&.162&.972&.016&.006&0&.248&.246&.224&.002&0&0&.066&.020&0\\
           30$\to$40&[28.51, 45.07]&[0.9541, 0.9980]&31.89/1.89&0&.200&.190&.966&.006&.012&0&.368&.492&.446&.010&0&0&.022&.012&0\\
            40$\to$50&[28.45, 44.46]&[0.9088, 0.9971]&41.37/1.37&.012&.082&.032&.892&.004&.112&0&.372&.578&.512&.008&.002&0&.012&.010&0\\
         \hline
    \end{tabular}}
    \label{Sec3::exp2}
\end{table*}

\begin{figure}[t]
\centerline{\includegraphics[width=1\columnwidth]{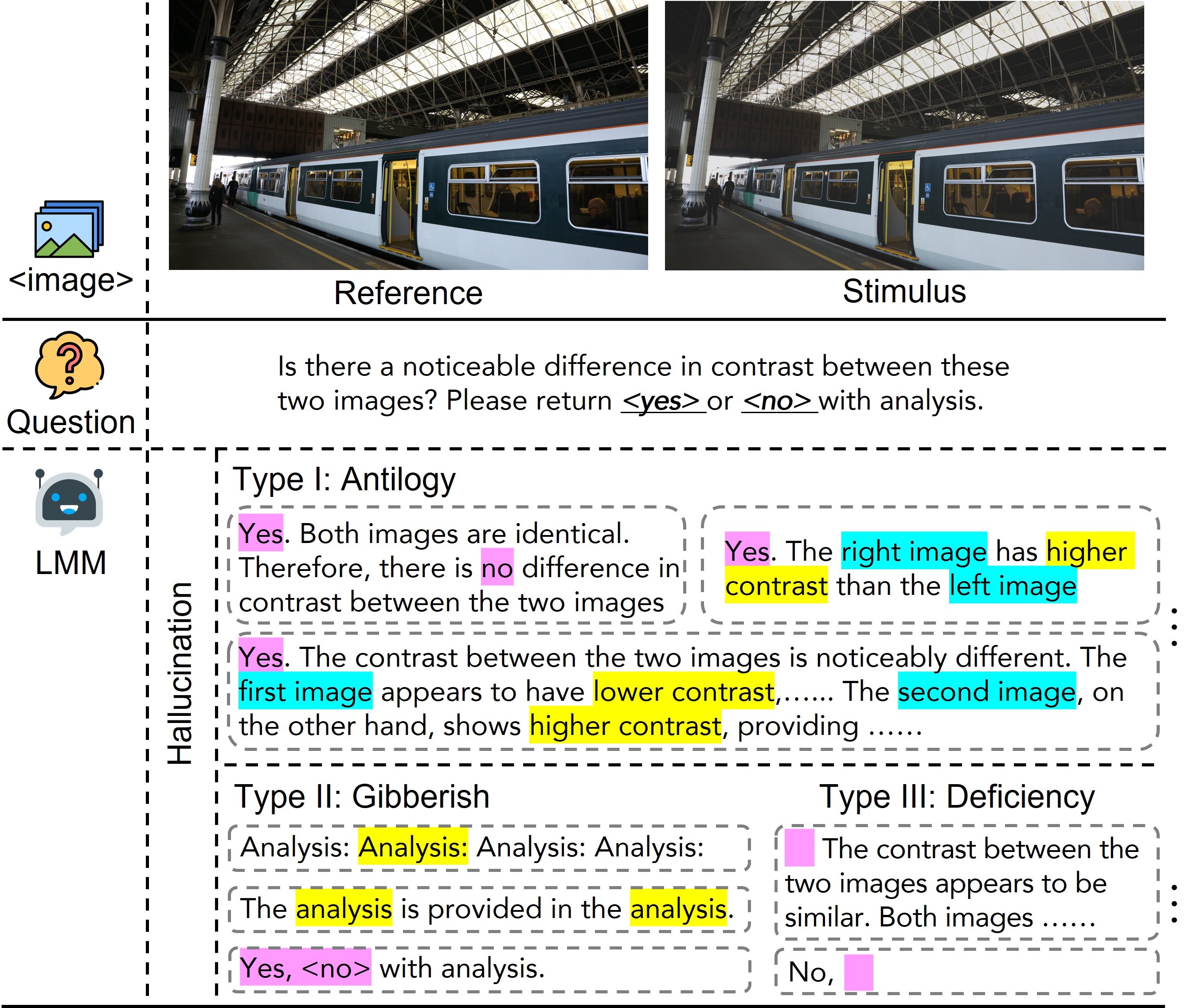}}
\caption{An example of hallucination in answers during the response sensitivity study for LMMs. We highlight the hallucination parts in different colors, where the output flag and analysis tokens show obvious antilogy, gibberish, or deficiency under the given question.}
\label{example-hallu}
\end{figure}

\section{Perceptual Blind Area of LMMs}
We first investigate the perceptual blind area of LMMs in terms of the vision and language perspectives. We select 5 representative LMMs series including Qwen2.5-VL \cite{bai2025qwen25vl}, DeepSeek-VL2 \cite{wu2024deepseek}, InternVL2.5 \cite{chen2024expanding}, Qwen2-VL \cite{wang2024qwen2}, and LLaVA-OneVision \cite{li2024llava}, as well as the currently smallest vision-language model, SmolVLM-256M \cite{SmolVLM2025} (Tab. \ref{tab:arch}), for these studies.
These LMMs not only vary in their vision architectures, such as QwenViT {\it vs.} SigLIP with different number of parameters (675M {\it vs.} 400M) and input sizes (224\textsuperscript{2} {\it vs.} 384\textsuperscript{2}), but also in their language backbones (Qwen2.5 {\it vs.} DeepSeekMoE) and scales (72B {\it vs.} 16B). Given their widespread use in various multimodal vision tasks and real-world applications, these LMMs provide a robust basis for an initial examination of perception defects.

\begin{figure}[t]
\setlength{\abovecaptionskip}{0.cm}
\centerline{\includegraphics[width=1\columnwidth]{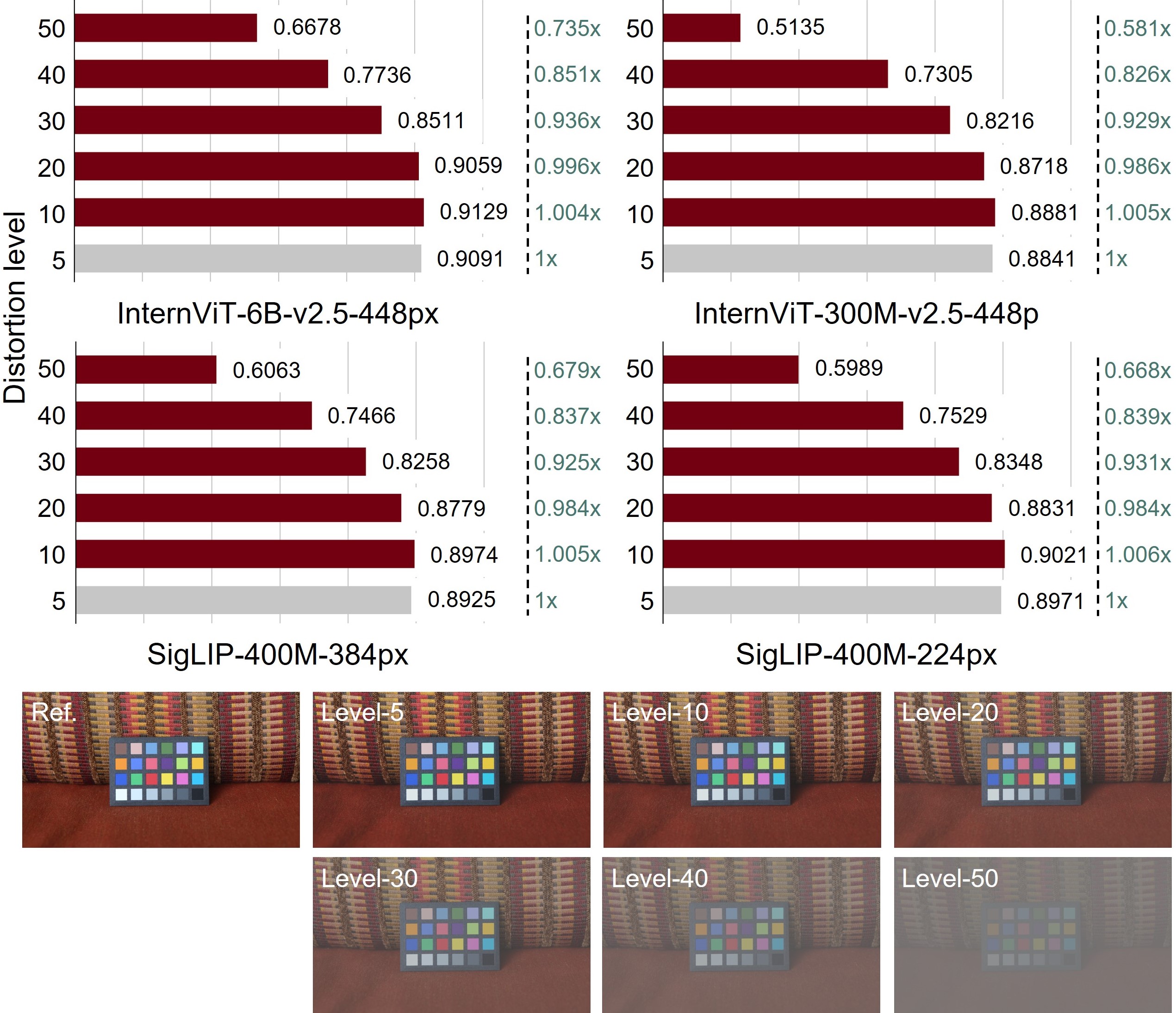}}
\caption{The cosine similarity between the original image embedding and
contrast-distorted image embedding among different visual encoders. Below the bar figure, we provide one visual example to illustrate the different distortion levels.}
\label{embedding-similarity}
\vspace{-2em}
\end{figure}

{\it In the first experiment, we demonstrate that different LMMs have similar visual shortcomings by comparing their perceptual changes under various distortion levels}. We distort 50 reference images from the MCL-JCI dataset \cite{jin2016statistical} with a common low-level contrast adjustment. Each image pixel is first normalized to [0, 1] and then processed by a sigmoid function:
\begin{equation}
    f(x) = 1/(1 + \exp ( - k*(x - 0.5)),
\label{contrast_eq}
\end{equation}
where $k$ is a scaling factor that controls the contrast levels. Specifically, we set $k$ to $[5:0.5:50]$ (which means values ranging from 5 to 0.5 with 50 levels). A lower $k$ indicates a lower contrast. 
Since many LMMs apply the same visual encoder, we choose four available vision backbones, {\it i.e.}, InternViT-6B-v2.5-448px, InternViT-300M-v2.5-448px, SigLIP-400M-384px, and SigLIP-400M-224px for comparison. 
For visual encoder $E_m(\cdot)$, we calculate a {\it visual perception correlation} $\rho$ on target image $I^{ref}_i$ across different distortion levels $k$, which can be formulated as:
\begin{equation}
   \rho = S\left( {{E_m}({I^{ref}_i}),{E_m}(I_i^k)} \right), 
\end{equation}
where $S( \cdot, \cdot )$ computes the cosine similarity between the original image embedding and the distorted image embedding. Fig. \ref{embedding-similarity} shows the average perception differences on six typical distortion levels $\{ 5,10,20,30,40,50\}$ for 50 tested images. We find that perceptual discrepancies do not proportionally amplify with an increasing level of distortion. Surprisingly, a level-10 contrast distortion exhibits higher perceptual similarity (1.004x\textendash1.006x) to the reference image compared to a level-5 distortion. Moreover, in the InternViT-6B visual encoder, even a level-20 distortion shows minor perceptual difference compared to level-5 distortion (0.996x), whereas the human eye can clearly observe visual differences at this level ({\it according to extra user studies on these images following the procedures in} \cite{lin2022large} {\it to obtain the minimum range of distortion (level $\approx5.83$) that humans can perceive}). This indicates that visual encoders in LMMs may have perceptual blind areas for low-level distortions. 

Furthermore, {\it are LMMs dumb in responding differently under various distortion levels? We notice a response commonality shared across existing LMMs, which manifests as disproportionate sensitivity to image distortions.} Here, we calculate a {\it response variation label} for the perception of LMM $M_m$ to $I_i^k$, which is denoted as:
\begin{equation}
    L\left({{M_m};{I_i}^{ref};I_i^k} \right) = \left\{ {\begin{array}{*{20}{c}}
  {1,}&{{\text{if}}\;{M_m}(I_i^{ref}, I_i^k) \cong Y^{flag}} \\ 
  {0,}&{{\text{otherwise}}} 
\end{array}} \right.,
\end{equation}
Where ${M_m}(I_i^{ref},I_i^k)={\hat Y^{flag}}+\mathcal{A}$. This label is 1 only if the model's joint perceptual response of the reference image and the distorted image is equivalent to the golden flag $Y^{flag}$. In this study, we use the `Yes-or-No' query form to determine whether the responses are equivalent by asking {\it `Is there any noticeable difference in contrast between the two images? Please answer {\tt $<$yes$>$} or {\tt $<$no$>$} with analysis'}. Note that the output flag ${\hat Y^{flag}}$ and analysis $\mathcal{A}$ should not contradict the meaning expressed by the golden flag, otherwise, it will be regarded as a hallucination. As shown in Fig. \ref{example-hallu}, we mainly encounter three types of hallucination, {\it i.e.}, antilogy, gibberish, and deficiency, during this experiment. Among them, antilogy primarily manifests as a semantic contradiction between the output flag and the analyses, which may appear plausible sounding but is actually a hallucination. Gibberish and deficiency represent meaningless and content-missing outputs, respectively, which occur more frequently in small models (Param$\leqslant$7B). 
After excluding the above types of hallucinations, we present the response sensitivity ({\it i.e.} frequency of label $L$ equivalent to 1) in Tab. \ref{Sec3::exp2}, where PSNR and SSIM variation ranges as well as the corresponding HVS-JND for different distortion intervals are also provided as references.
First, the average response sensitivity scores of all tested LMMs across six contrast distortion levels are about 0.091, 0.096, 0.110, 0.135, 0.170, and 0.164, respectively. This indicates that LMMs exhibit varying perceptual sensitivity across different distortion levels, with higher sensitivity observed under more severe distortions (level $>30$). Such a finding is similar to the human perception mechanism. Second, smaller models are less responsive to distortion changes, such as SmolVLM-256M, DeepSeek-VL2-Tiny, Qwen2-VL-2B, and LLaVA-OneVision-0.5B, which are almost all lower than 1e-4. Surprisingly, Qwen2.5-VL-3B exhibits extremely high response sensitivity (avg. $=0.887$) compared to models of comparable scale, even surpassing those that are an order of magnitude larger. Meanwhile, we observe that the response sensitivity is not necessarily proportional to the model scale. Specifically, the highest score of Qwen2.5-VL, DeepSeek-VL2, and InternVL2.5 series is found in Qwen2.5-VL-3B, DeepSeek-VL2-Small, InternVL2.5-38B, respectively. 
Third, We notice that the InternVL2.5 series covers a wider contrast distortion response range than other LMM series. 
In conclusion, we should carefully consider such perceptual granularity discrepancies among LMMs.

\section{LMM-Oriented JND}
To systematically study the perceptual granularity of LMM, we propose the concept of LMM-oriented just noticeable difference, {\it i.e.}, LMM-JND. In this section, we first introduce our motivation for this study in Sec. \ref{Formulation}, and then elaborate on the definition of LMM-JND as well as its determination pipeline in Sec. \ref{Determination}.

\subsection{Motivation and Task Formulation} 
\label{Formulation}
The HVS-JND refers to the visual redundancy of images or videos, technically defined as the imperceptible threshold of each pixel or the associated sub-bands, which is usually collected by conducting large-scale subjective experiments \cite{lin2021progress}. Hence, HVS-JND is a representation of population perception ability. Differing from HVS-JND, where the human eye serves as the ultimate receptor, the proposed LMM-JND aligns with the modality types processed by LMMs. In this work, we first aim to identify the minimal perceptible quantities for LMMs, {\it i.e.} the $1$\textsuperscript{st} JND, and then investigate whether LMMs exhibit special perceptual redundancy characteristics similar to those of the HVS (stair-like $i$\textsuperscript{th} JND). Any changes under the LMM-JND magnitude will not affect the response of the LMM.
Although existing LMMs exhibit significant variations in performance across different tasks \cite{wu2023multimodal,yin2024survey}, their main techniques, {\it e.g.}, vision encoder, language backbone, adapter, and scale, that they used are similar, which suggests that they may have identical LMM-JND characteristics. 
Therefore, we use the proposed VPA-JND dataset containing diverse visual stimuli to probe LMM-JND in a more reasonable and generalized manner.

\begin{figure}[t]
\centerline{\includegraphics[width=\columnwidth]{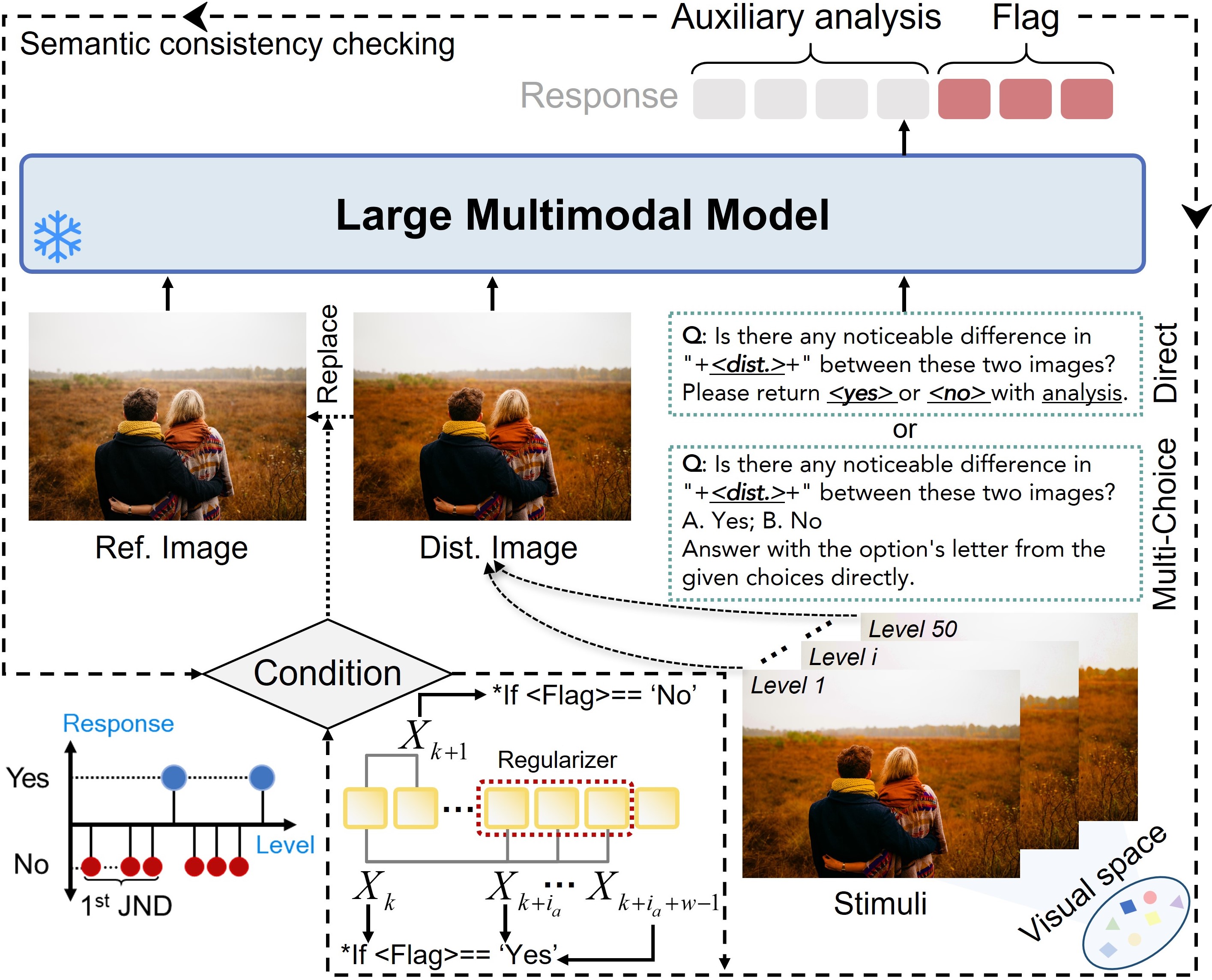}}
\caption{The overall determination pipeline of LMM-JND. To be eligible for the LMM-JND study, an LMM should inherently support multi-image input to achieve the best performance.}
\label{pipeline}
\end{figure}

{\bf Formulation.} We define LMM-JND as the minimal amount of stimulus $X$ that must be changed for an input difference $\Delta X$ to be processed by an LMM, where $X$ can be signals from any modality\footnote{In this paper, we mainly discuss the visual signals.}, depending on the LMM itself. Then, the LMM-JND can be generally formulated as:
\begin{equation}
   \text{LMM-JND}\left( X \right) = \Phi \left( {{\theta _1},{\theta _2}, \cdots ,{\theta _k}}; X \right) + \Delta {X_l},
\end{equation}
where $\Phi \left(  \cdot ;X \right)$ indicates a function of multi-parameter combination, which reflects the variation tendency or tolerance of $\text{LMM-JND}\left( X \right)$ under different stimuli $X$. It can be constant, linear, exponential, logarithmic, {\it etc}. 
The value of $k$ depends upon the input modality, while $\Delta {X_l}$ denotes the spontaneous terms in the absence of external stimuli, {\it i.e.}, the perceptual limit inherent in the network architecture. Furthermore, since LMM-JND deals with the non-distinguishability of LMMs, it is non-transitive. In other words, if response ${\mathrm{M}\left(X_0, X_1\right)} \cong \mathrm{M}\left(X_1, X_2\right)$, where $\mathrm{M}\left(\cdot,\cdot\right)$ denotes the multi-image input of LMMs, $X_0$ is still possible to distinguish from $X_2$ (denoted as ${X_0} \ncong {X_2}$). This phenomenon illustrates the necessity of employing sequential paired comparisons in LMM-JND evaluations.

\subsection{Single and Multiple LMM-JND Determination}
\label{Determination} 
In Fig. \ref{pipeline}, we illustrate the overall determination pipeline for LMM-JND.
Intuitively, given a reference input $X_0$, the stimulus at $k$-th stage can be denoted as:
\begin{equation}
    {X_k} = {X_0} + {v_k},k = 1,2, \ldots
\end{equation}
where $v_k$ denotes the $k$-th signal variation associated with the generation of stimuli, which can be any modality supported by the LMM. 
Then, if ${\mathrm{M}\left(X_0, X_k\right)} \cong {\mathrm{M}\left(X_0, X_{k-1}\right)}$ holds for $k = 2, \ldots ,N - 1$, but ceases to be true for $k \geq N$, then the LMM-JND is defined as:
\begin{equation}
    {\text{LMM-JND}}\left( X \right) = \left\{ {\begin{array}{*{20}{c}}
  {{X_N} - {X_0},}&{{\text{signal form}}} \\ 
  {N,}&{{\text{quantization level}}} 
\end{array}} \right.
\end{equation}
This is shown in Fig. \ref{pipeline}, as the process to locate the $1$\textsuperscript{st} JND, that is, the distortion level when the condition is met for the first time.
An iterative paired comparison procedure is then applied to find all possible JNDs for a stimulus as follows.
\begin{compactitem}
    \item {\bf Step-1:} For finding the $1$\textsuperscript{st} JND, let the index of the anchor to start from zero, {\it i.e.}, set the original reference image $X_{k=0}$ as {\it anchor 0}.
    \item {\bf Step-2:} Applying either direct question-answering or multi-choice questions prompting for comparing image pairs $({X_k},{X_{k + 1}}),({X_k},{X_{k + 2}}), \ldots ,({X_k},{X_{k + i}}), \ldots ,$ until $({X_k},{X_{k + {i_a}}})$, which is the first input image pair judged as having noticeable difference, or mathematically, ${\mathrm{M}\left(X_k, X_{k+{i_a}-1}\right)} \ncong \mathrm{M}\left(X_k,X_{k+{i_a}}\right)$. Hence, $X_{k+{i_a}}$ is the stimulus with a LMM-JND.
    \item {\bf Step-3:} Afterwards, the current anchor index $k$ is updated to $k+{i_a}$, namely the reference image is $X_{k+{i_a}}$, and repeating {\bf Step-2} until all LMM-JNDs are found.
\end{compactitem}
However, due to the instability of outputs from LMMs, there may exist a tendency for the LMM to prefer certain types of responses, leading to unreliable judgments for the JND points. For example, a false LMM-JND determination will happen when the preceding items and ${\mathrm{M}\left(X_k, X_{k+{i_a}-1}\right)} \ncong \mathrm{M}\left(X_k,X_{k+{i_a}}\right)$ are all satisfied, while ${\mathrm{M}\left(X_k, X_{k+{i_a}-1}\right)} \cong \mathrm{M}\left(X_k,X_{k+{i_a+1}}\right)$.
To avoid this problem, we introduce a {\it sliding window regularizer} for additional scrutiny. As illustrated in Fig. \ref{pipeline}, we add extra response comparisons after the potential LMM-JND point within a window of width $w$ ({\it i.e.} the red dotted rectangle end with $X_{k+{i_a}+w-1}$). The next iteration will proceed only if all stimuli within the window meet the specified conditions.

\begin{algorithm}[t]
\caption{Pseudocode for LMM-JND Determination} 
\label{alg:example}
\begin{algorithmic}
\STATE {\bfseries Input:} An LMM $\mathrm{M}$, NLI model $\mathrm{NLI}$, reference image $X_0$ and its stimulus set $\{X_1,\dots,X_N\}$, pre-defined flag $f_{+}$, ground-truth query-answer pair $G, Q$, window width $w$.
\STATE {\bfseries Output:} An LMM-JND list $L$ for stimulus $X$. 
\STATE {\it Initialize} reference anchor $k = 0$ and L=[ ]% , order $t=1$
\FOR{$k=0$ to $N-1$}
\FOR{$i=1$ to $N-k$}
\STATE $R_f,R_{aux}={\mathrm{M}}\left( {{X_k},{X_{k + i}}}\right)$
\IF{$f_{+} \notin R_f $ or $\mathrm{NLI}\left( {{R_{aux}},G|Q} \right)=False$}
\STATE continue
\ENDIF
\STATE $cnt=0$
\FOR{$n=1$ to $w-1$}
\STATE $R_f,R_{aux}={\mathrm{M}}\left( {{X_k},{X_{k + i+n}}}\right)$
\IF{$f_{+} \in R_f $ and $\mathrm{NLI}\left( {{R_{aux}},G|Q} \right)=True$}
\STATE $cnt=cnt+1$
\ELSE
\STATE  break 
\ENDIF
\ENDFOR
\IF{$cnt=w-1$}
\STATE $k=k+i$
\STATE $L$\texttt{.append(k)}
\STATE break
\ENDIF
\ENDFOR
\ENDFOR
\STATE {\bfseries return} LMM-JND list $L$ 
\end{algorithmic}
\end{algorithm}

\begin{table*}[t]
    \centering
    \renewcommand\arraystretch{1.15}
    \caption{The Average Maximum and Minimum Distortions across Different Dimensions in the VPA-JND, Characterized by PSNR, SSIM, and LPIPS Metrics}
    \resizebox{1\linewidth}{!}{\begin{tabular}{c|cccccc|ccc|cc}
    \hline                 
        {Metric}  & {\textit{Blur}}& {\textit{Brightness}} & {\textit{Color}} & {\textit{Contrast}} & {\textit{JPEG}} & \textit{Noise} & \textit{Mask} &\textit{Watermark\textsubscript{qrcode}}&\textit{Watermark\textsubscript{text}}&\textit{Angle}&\textit{Distance}  \\ \hline
         PSNR&[21.43, 32.05]&[7.37, 56.91]&[16.09, 62.27]&[11.72, 30.41]&[22.37, 42.09]&[15.35, 47.02]&[21.22, 51.51]&[23.31, 58.85]&[38.33, 72.19]&[8.33, 17.59]&[14.33, 26.88]\\
         SSIM&[0.6496, 0.9200]&[0.5735, 0.9992]&[0.8544, 0.9996]&[0.5899, 0.9521]&[0.6662, 0.9992]&[0.2164, 0.9942]&[0.9614, 0.9997]&[0.9433, 0.9995]&[0.9916, 0.9999]&[0.3634, 0.5674]&[0.4860, 0.6038]\\
        LPIPS&[0.1481, 0.6066]&[0.0003, 0.5278]&[0.0001, 0.2635]&[0.0347, 0.5293]&[0.0035, 0.8574]&[0.0038, 0.9934]&[0.0048, 0.0536]&[0.0005, 0.0813]&[$<$10e-4, 0.0270]&[0.3634, 0.5675]&[0.2732, 0.5998]\\
        \hline
    \end{tabular}}
    \label{dataset_diversity}
\end{table*}

Furthermore, considering the response characteristics of LMMs \cite{wu2024qbench,chen2024obi}, the LMM-JND determination task is suited to either a flag-guided generalized answer or a multiple-choice answer format. 
While mitigating the impact of hallucinations shown in Fig. \ref{example-hallu} is also essential to ensure the reliability of the LMM-JND determination process.
Specifically, we design a ground truth term $G$ to check the semantic consistency of the output flag and context:
\begin{equation}
    G = \mathbb{I}\left( {{f_{+}\in R_f}} \right) \cdot {G^P} + \mathbb{I}\left( {{f_{-}}\in R_f} \right) \cdot {G^N}
\end{equation}
where $\mathbb{I}(\cdot)$ denotes the indicator function that equals $1$ if the condition is true and $0$ otherwise. $f_{+}$ and $f_{-}$ are the pre-defined flag and its negative version. $G^P$ and $G^N$ denote the corresponding ground truth positive and negative answers respectively. $R_f$ represents the flag tokens in the response. 
Then, we introduce a new metric, {\it Minimum Response Variation} ($\mathcal{MRV}$), to quantify the minimal perceptible distortion level of an LMM at any $n$\textsuperscript{th} JND. Since a single type of stimulus-response only considers the visual granularity in a narrow scope, $\mathcal{MRV}$ averages the response variations across a range of stimuli.
Specifically, for a `Yes-or-No' question $Q$, given a stimulus type $s$ and pre-defined flag ($f_{+}$=`{\it yes}'), $\mathcal{MRV}$ at $1$\textsuperscript{st} JND can be formulated as:
\begin{equation}
    \mathcal{M}\mathcal{R}{\mathcal{V}_s^{1}} = \frac{1}{T}\sum\limits_{j=1}^T {\min \left\{ {k|{f_ + } \in {R_f} \wedge \mathrm{NLI}\left( {{R_{aux}},G|Q} \right)} \right\}} ,
\end{equation}
where ${R_f}$ and ${R_{aux}}$ represent the flag term and the auxiliary analysis term in the output response $\mathrm{M}_s( {X_0^j,X_k^j})$ respectively.
$T$ is the total number of references. 
$X^{j}_{k}$ denotes the $j$-th stimulus at $k$-th distortion level. 
$\mathrm{NLI}(\cdot)$ refers to a natural language inference (NLI) model, which is employed to detect semantic contradictions.
The corresponding pseudo-code is provided in Algorithm \ref{alg:example}.

Overall, this approach allows for the study of LMM-JND by collecting large-scale multimedia JND datasets and offers a discriminative measurement for comparing the minimal perceptible details across different LMMs.
LMMs with lower $\mathcal{MRV}$ are more sensitive to subtle input variations, which indicates superior perceptual acuity but may compromise robustness in specific scenarios ({\it e.g. anti-disturbance when encountering adversarial attacks}).
More importantly, based on LMM-JND, we can identify their perceptual characteristics and commonality curves, thereby guiding the elimination of perceptual redundancy in LMMs.
Therefore, the study of LMM-JND is an essential research direction in the current era and plays a critical role in both the offensive and defensive aspects of LMMs.

\section{VPA-JND Dataset}
\begin{figure*}[t]
\begin{center}
\includegraphics[width=1\linewidth]{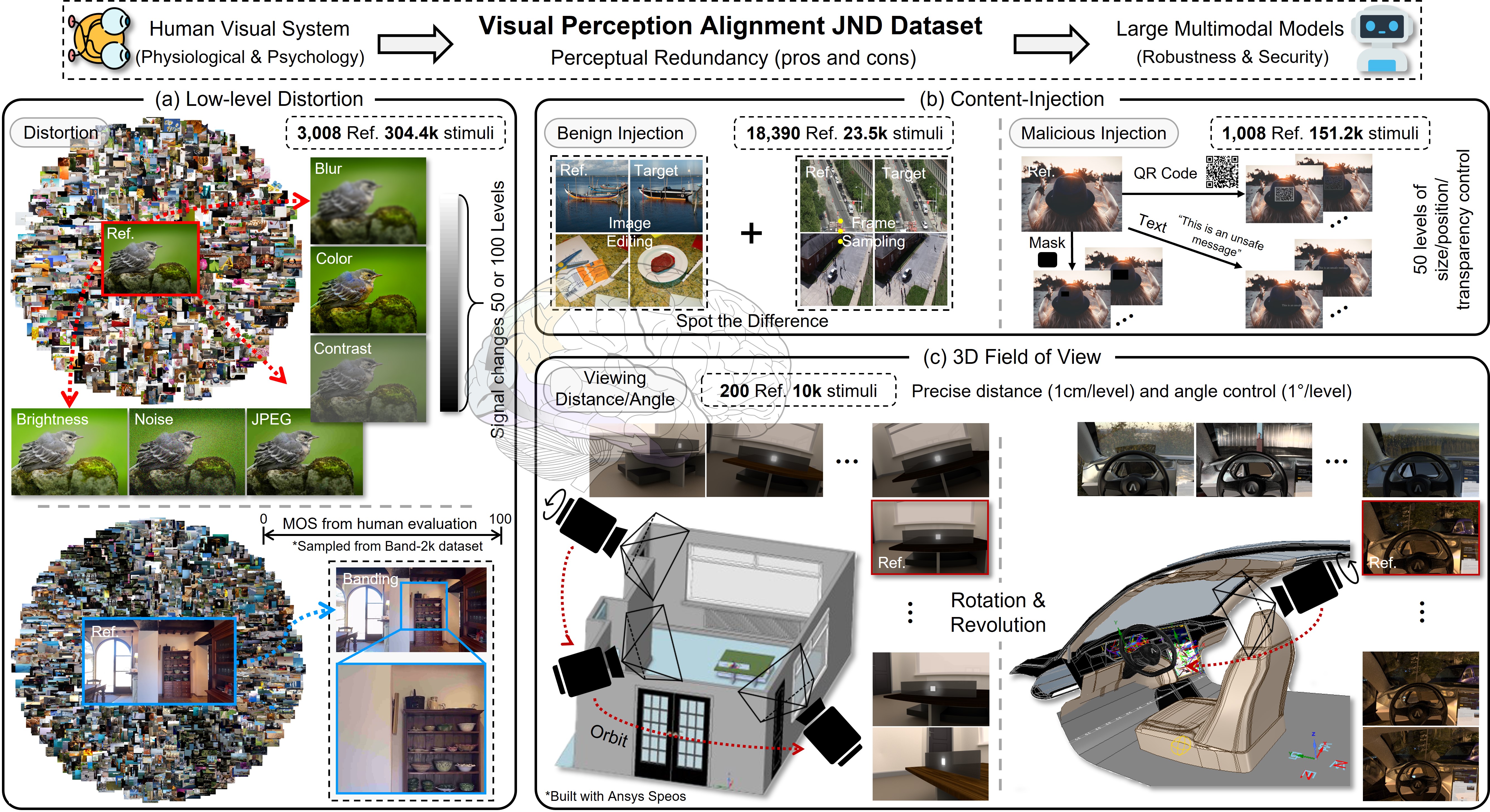}
\caption{Overview of the proposed visual perception alignment JND dataset ({\bf VPA-JND}). It comprises three parts: (a) 3,008 reference images and 304.4k stimuli with seven types of low-level distortions ({\it i.e.}, {\it blur}, {\it color}, {\it brightness}, {\it noise}, {\it contrast}, {\it JPEG}, and {\it banding artifacts}); (b) A total of 19,398 reference images and 174.7k stimuli that involve two types of content-injections, {\it i.e.}, {\it benign} and {\it malicious}; (c) 200 reference images and 10k spatial-aware stimuli obtained by changing the 3D field of view (FoV) in different simulation environments.}
\label{VPA-JND}
\end{center}
\end{figure*}

In this section, we establish the first JND dataset for LMMs, dubbed the visual perception alignment JND dataset (VPA-JND), shown in Fig. \ref{VPA-JND}, as a prerequisite for assessing the perceptual granularity of LMMs. Our VPA-JND dataset has the following notable advantages:
\begin{itemize}
    \item {\it Large-scale}: VPA-JND is the first and so far the largest dataset dedicated to evaluating the perceptual granularity of LMMs, which contains 21,598 reference images and 489,065 stimuli.
    \item {\it HVS-aligned}: The design of the VPA-JND dataset draws inspiration from the signal processing mechanisms of the human visual cortex ({\it v1-v4}) \cite{grill2004human}, which strategically aligns with the human-like development objectives of contemporary LMMs.
    \item {\it Comprehensiveness:} The proposed VPA-JND dataset involves 7 common low-level distortions and 2 content-injection patterns associated with 5 visual understanding tasks, as well as 2 major spatial FoV changes, which comprehensively approach real applications.
\end{itemize}

\subsection{Source Content and Stimulus Generation}
\label{dataset}
Neuroscientific evidence \cite{grill2004human,bitar2016algorithmic,ficsek2023cortico} suggests that ocular sensory input is initially processed in the primary visual cortex ({\it v1}), where low-level perceptual features ({\it e.g.}, blurriness, edge orientation, and spatial frequency) undergo preliminary encoding. Then, {\it v1} transmits information to two pathways, {\it i.e.}, the ventral stream and dorsal stream \cite{goodale1992separate,milner2017two}, to handle high-level visual perception. The ventral stream is associated with object recognition and form representation, while the dorsal stream is involved in spatial information awareness.
Aligning with this, our VPA-JND dataset comprises 3 subsets: low-level distorted images, content-injected images, and 3D field of view (FoV) images. 
The collected stimuli exhibit various properties, including various resolutions (224\textsuperscript{2}-1080p), different scenarios, and tampering means, thus ensuring the diversity of our dataset.
Fig. \ref{VPA-JND} shows some examples in our VPA-JND dataset. 
In total, we collect 489k stimuli from 21k reference images encompassing the following types:

{\bf Low-Level Distortions.} We consider 7 typical low-level distortions, {\it i.e.}, blur, brightness, color saturation, contrast change, JPEG compression, and banding artifacts, resulting from the signal acquisition, transmission, or quantization that deal with efficient visual coding \cite{lin2019kadid,chen2025joint}. 
To ensure the diversity of the reference image, we utilize 1,008 images from the KonJND-1k reference set \cite{lin2022large} to generate stimuli.
Specifically, for blur, we employ a Gaussian blur kernel with the blur radius set to $[1:10:50]$ (which represents values ranging from 1 to 10 with 50 equidistant levels). For brightness, we first convert the image from RGB space to HLS space and then scale the luminance component by a multiplicative factor set to $[1:0.1:50]$. For color distortion, we convert the image from RGB space to HSV space and scale the saturation component by a factor set to $[1:5:50]$. For contrast changes, we apply the same strategy in Eq. \ref{contrast_eq}. For noise, we generate a set of Gaussian white noise by setting the variance to $[1:50:50]$ and add it to the original images.
For JPEG compression, we have distortion levels $d=101-QF$, where $QF$ denotes the JPEG quality factor set to $[100:1:100]$, and only half of the reference images are selected to maintain equilibrium in the overall sample population.
Since banding artifacts cannot be generated with controllable intensity parameters, we use the rescaled human-evaluated mean opinion scores (\text{MOS}) $\in [0,100]$ from BAND-2k dataset \cite{chen2024band} as distortion levels. 
Consequently, a total of 304.4k stimuli are obtained by corrupting 3,008 reference images with distortions at different degradation levels.

\begin{figure}[t]
\setlength{\abovecaptionskip}{0.cm}
\centerline{\includegraphics[width=1\columnwidth]{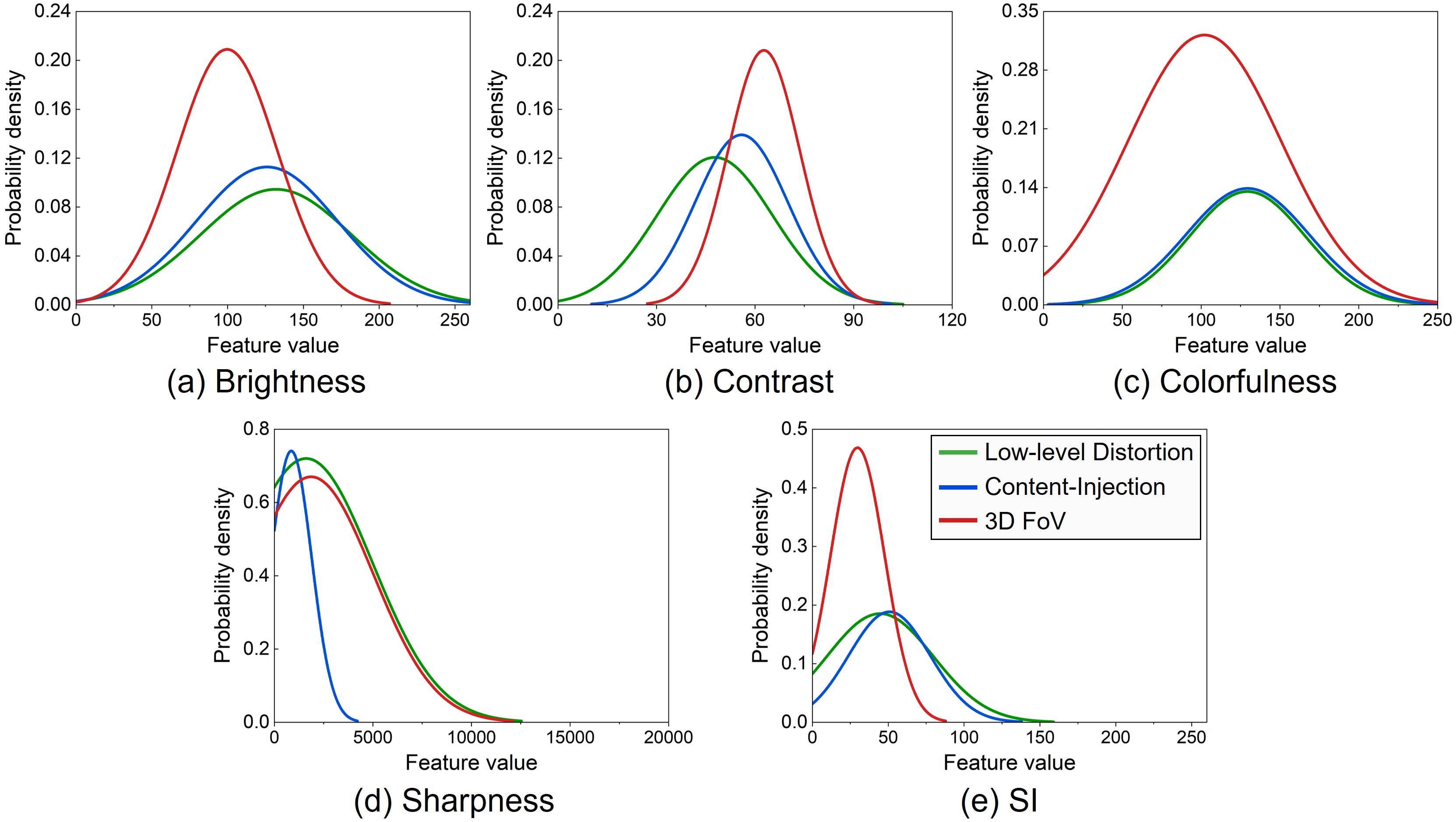}}
\caption{Feature distribution comparisons among three considered evaluation perspectives: low-level distortion, content-injection, and 3D FoV.}
\label{feature_distribution}
\end{figure}

{\bf Content-Injection.} Grounded in visual acuity tests and robustness benchmarks \cite{carlson2016clinical,tong2024eyes,cui2024robustness} as well as the consideration of the input security for LMMs, we outline two different content perturbations, {\it i.e.}, benign and malicious, that require qualitative and quantitative perception of semantic content, location, and visibility. Concretely, we treat benign injection as a spot-the-difference task. For this purpose, an image editing dataset, MagicBrush \cite{zhang2023magicbrush}, and a visual difference dataset \cite{jhamtani2018learning} collected by frame sampling are chosen as the evaluation set. 
On the other hand, we use a scalable mask with variable size and position and two types of transparency-controlled watermarks ({\it i.e.}, QR code and text) to achieve malicious content injection, which presents substantial challenges to the security and legal compliance of input content in LMMs. The mask size is set to $[1\%:25\%:50]$ of the reference image dimensions 
(cf. Weber-Fechner law in psychophysics \cite{weber1996eh}). Similarly, we control the transparency of the injected watermark to $[1\%:50\%:50]$. In this part, 1,008 reference images from the KonJND-1k dataset are used for stimuli generation. Eventually, we obtained 19,398 reference images with over 174.7k stimuli for the content-injection set.

\begin{table*}[t]
    \centering
    \renewcommand\arraystretch{1.15}
    \caption{The Minimal Perceptible Distortion Level ($1$\textsuperscript{st} JND) of Various LMMs on Different Visual Stimuli. `Avg.' is the Average of All $\mathcal{MRV}_{s\in \mathcal{S}}$ Sums for the Stimulus Set $\mathcal{S}$. The General Maximum Value for each Cell is 50, Except for the {\it JPEG} and {\it Banding} Columns, where the Maximum Value is 100. `WM' is the Abbreviation for the Embedded Watermark. The Best and Second-Best Results are Highlighted in {\bf Bold} and \underline{Underlined}, Respectively}
    %\vspace{-8pt}
    \resizebox{1\linewidth}{!}{\begin{tabular}{r|ccccccc|ccc|cc|c|c}
    \hline                 
        \textbf{Test Set} & \multicolumn{7}{c|}{\textbf{Low-level Distortion}} &\multicolumn{3}{c|}{{\bf Content-Injection}}& \multicolumn{2}{c|}{\textbf{3D FoV}}&\multirow{2}{*}{Avg.}&\multirow{2}{*}{Rank} \\ \cdashline{1-13}
        {\textbf{Model} (Index)}  & {\textit{Blur}}& {\textit{Bri.}} & {\textit{Color}} & {\textit{Contr.}} & {\textit{JPEG}} & \textit{Noi.}  & \textit{Band.} &\textit{Mask}&\textit{WM\textsubscript{qrcode}}&\textit{WM\textsubscript{text}}&\textit{Angle}&\textit{Distance}&  \\ \hline
\rowcolor{cyan!10} \textsc{Human Level}&1.24&8.91&4.18&3.42&52.68&2.24&\textendash&1.02&1.56&1.73&1.13&1.83&\textendash&\textendash\\
        \hline
         \multicolumn{10}{l}{\textit{Open-source LMMs}} \\ \hdashline
         SmolVLM-256M (A)&38.65&44.93&43.26&46.36&99.34&43.79&8.32&42.73&39.33&49.91&4.75&5.67&38.92&16\\
         Qwen2.5-VL-72B (B)&2.87&32.56&25.46&29.42&97.19&12.83&\underline{2.26}&11.74&11.76&29.63&5.50&26.33&23.96&6\\
         Qwen2.5-VL-7B (C)&2.79&32.00&22.16&25.72&98.23&15.70&2.88&42.73&25.11&40.95&6.50&33.50&29.02&11\\
         Qwen2.5-VL-3B (D)&2.37&40.95&36.99&36.39&96.58&28.01&4.32&47.50&13.67&29.89&9.00&12.33&29.83&12\\
         DeepSeek-VL2 (E)&2.60&43.05&32.02&48.37&98.18&23.63&5.45&14.01&26.75&42.55&17.75&49.38&33.65&14\\
         DeepSeek-VL2-Small (F)&\underline{1.27}&33.18&27.56&30.39&95.49&9.27&7.50&49.87&49.88&49.97&13.50&18.83&32.23&13\\
         DeepSeek-VL2-Tiny (G)&$>$50&$>$50&46.67&$>$50&$>$100&49.96&$>$100&$>$50&49.97&$>$50&$>$50&$>$50&$>$58.05&20\\
         InternVL2.5-78B (H) &1.77&21.84&17.46&20.43&46.43&17.48&{\bf 1.23}&28.33&14.28&28.72&7.17&21.33&18.87&4\\
         InternVL2.5-38B (I)&1.87&\underline{19.82}&13.53&22.79&52.09&16.34&2.57&22.68&{\bf 10.37}&\underline{21.14}&7.67&23.74&17.88&\colorbox{green!20}{3}\\
         InternVL2.5-8B (J)&2.07&27.38&21.10&26.22&76.85&23.19&2.46&27.31&16.71&25.11&11.25&29.17&24.07&8\\
        Qwen2-VL-72B (K)&4.78&31.11&26.39&33.21&98.55&{\bf 7.29}&4.93&17.45&13.41&36.06&8.03&33.39&26.22&10\\
         Qwen2-VL-7B (L)&1.53&24.96&31.93&26.63&99.52&10.62&26.82&48.82&43.43&48.30&10.25&34.50&33.94&15\\
         Qwen2-VL-2B (M)&42.63&34.49&40.23&43.63&99.54&34.68&4.41&48.02&48.89&49.49&26.84&40.33&42.77&17\\
         LLaVA-OneVision-72B (N)&2.86&33.35&22.56&34.68&85.75&7.88&3.60&16.57&13.28&35.67&7.50&24.33&24.00&7\\
        LLaVA-OneVision-7B (O)&47.69&48.31&38.59&47.07&99.64&49.13&$>$100&48.92&46.53&48.94&47.33&39.17&$>$55.11&18\\
         LLaVA-OneVision-0.5B (P)&49.68&49.09&49.87&49.23&99.65&49.65&$>$100&42.38&46.70&33.76&48.25&47.00&$>$55.44&19\\
        \hline
        \multicolumn{10}{l}{\textit{Proprietary LMMs (API)}} \\ \hdashline
         \textsuperscript{{\it ver. 1120}}GPT-4o (Q)&{\bf 1.25}&20.26&\underline{9.69}&{\bf 14.96}&{\bf 41.28}&\underline{7.76}&8.32&\underline{8.94}&14.17&25.15&\underline{1.50}&\underline{4.83}&\underline{13.18}&\colorbox{green!40}{2}\\
        Gemini 2.0 Flash (R)&\underline{1.27}&{\bf 19.66}&{\bf 8.93}&\underline{15.57}&\underline{44.38}&8.69&9.87&{\bf 8.38}&\underline{10.58}&{\bf 20.37}&{\bf 1.33}&{\bf 3.67}&{\bf 12.73}&\colorbox{green}{1}\\
        Gemini 1.5 Pro (S)&1.31&32.82&20.86&25.88&89.82&10.32&12.23&11.32&14.63&24.99&3.25&14.67&21.84&5\\
         \textsuperscript{{\it ver. 1119}}Qwen-VL-Max (T)&1.41&33.52&21.85&32.06&82.56&11.78&16.96&9.87&17.68&26.47&13.25&33.67&25.07&9\\
         \hline
    \end{tabular}}
    \label{exp1}
\end{table*}

{\bf 3D Field of View.} Visual-spatial observations are the basics of perceiving, understanding, and interacting in a sensory-rich 3D world. At present, the spatial intelligence of LMMs is a topic of widespread discussion \cite{yang2024thinking,qi2025shapellm}, with accurate first-person angle and distance perception serving as the foundation for spatial object localization.
To quantify such perception granularity, we build two virtual 3D environments using an optical simulation software, Ansys Speos \cite{ansys}, which enables us to achieve precise and controllable camera FoV adjustment.
We mainly focus on the rotation and revolution of the camera and select 200 reference viewpoints in space. Taking 1cm and 1\textsuperscript{${\circ}$} as the minimum step increments, we generate 10k stimuli with varying viewing distances and angles, including panning, zooming in, horizontal flipping, and pitch transformations.

\subsection{Dataset Analysis}
To validate the diversity of our proposed VPA-JND dataset, we analyze the image quality-related attributes. Specifically, we follow \cite{chen2025study} to employ 5 low-level features, including brightness, contrast, colorfulness, sharpness, and spatial information (SI), thereby providing a large visual space in which to depict and analyze statistical characteristics of our VPA-JND. Fig. \ref{feature_distribution} shows the fitted distribution curves of each selected feature. It can be observed that the low-level distortion set and content-injection set exhibit similar coverage in terms of brightness, colorfulness, and SI, while the 3D FoV set adheres closer to small values, which is consistent with the source contents in respective sets.
Overall, all features show a relatively Gaussian-like distribution with a single peak.
Moreover, we report the average maximum and minimum distortions across 11 evaluated dimensions in terms of three full-reference IQA metrics, including PSNR, SSIM \cite{wang2004image}, and LPIPS \cite{zhang2018unreasonable}, in Tab. \ref{dataset_diversity} for reference. 
Among them, the content-injection set owns a relatively small range of SSIM and LPIPS, which can be attributed to the fact that most of its disturbances are localized structural perturbations. Regarding 3D FoV set, the absolute values of all three metrics are inferior to those of the other two sets. This is because the 3D scene simulation output based on Speos not only alters the spatial viewing orientation but also modifies environmental information such as lighting and shadows.
Note that, based on the findings in \cite{tu2020bband,chen2024band,chen2024fs}, the PSNR, SSIM, and LPIPS metrics exhibit relatively low correlation with the severity of banding artifacts. Therefore, we omit these items.

\section{Experiments}

\subsection{Experimental Settings}

{\bf LMM Diversity.} In addition to the 16 open-source models listed in Tab. \ref{tab:arch}, we also include 4 mainstream proprietary LMMs, {\it i.e.}, GPT-4o (\texttt{2024-11-20}) \cite{GPT-4o}, Gemini (2.0 Flash \cite{gemini2} and 1.5 Pro \cite{reid2024gemini}), and Qwen-VL-Max (\texttt{2024-11-19}) \cite{bai2023qwen} for evaluation. All models natively support multi-image inputs to ensure the best performance.

\begin{table*}[t]
    \centering
    \renewcommand\arraystretch{1.25}
    \caption{The Equivalent PSNR and SSIM of the Minimal Perceptible Distortion Level for LMMs and Humans with Slash Separated. Note that the Rank of this Table may not be Fully Aligned with Tab. \ref{exp1}, since the JND Points Distribution Varies among Different Models}
    \resizebox{1\linewidth}{!}{\begin{tabular}{c|cccccc|ccc|cc}
    \hline                 
        {\textbf{Ind.}}  & {\textit{Blur}}& {\textit{Brightness}} & {\textit{Color}} & {\textit{Contrast}} & {\textit{JPEG}} & \textit{Noise} & \textit{Mask} &\textit{WM\textsubscript{qrcode}}&\textit{WM\textsubscript{text}}&\textit{Angle}&\textit{Distance}  \\ \hline
        \rowcolor{cyan!10} Hum.&32.01/.9149&27.63/.9834&31.24/.9913&30.43/.9503&32.81/.9433&41.22/.9792&51.50/.9997&31.56/.9946&68.19/.9998&17.44/.5646&25.46/.5948\\
        \hline
         A&22.19/.6627&8.96/.6548&16.51/.8659&12.58/.6502&22.37/.6664&16.39/.2495&22.79/.9714&25.41/.9476&38.34/.9916&13.93/.5053&21.95/.5732\\
         B&29.71/.8684&13.25/.7989&18.53/.9107&17.66/.8463&23.48/.7135&26.31/.6568&34.56/.9968&35.94/.9748&42.85/.9941&13.34/.4963&16.55/.5193\\
    C&29.88/.8713&13.43/.8030&19.12/.9201&19.18/.8737&22.48/.6775&24.51/.5814&22.78/.9716&29.33/.9568&40.11/.9926&12.61/.4855&15.91/.5102\\
       D &30.42/.8854&10.26/.7059&17.04/.8793&15.36/.7823&24.18/.7381&19.90/.3782&21.83/.9652&34.66/.9711&42.81/.9940&12.28/.4609&19.43/.5458\\
         E &30.07/.8767&9.57/.6793&17.57/.8914&12.33/.6317&22.54/.6817&21.28/.4370&33.18/.9960&28.78/.9556&39.79/.9923&10.21/.4068&14.41/.4817\\
        F&\underline{31.94/.9143}&13.01/.7906&18.14/.9031&17.37/.8389&25.13/.7626&29.11/.7652&21.32/.9618&23.33/.9435&38.34/.9916&10.95/.4310&18.02/.5342\\
         G &$<$21.43/.6496&$<$7.37/.5735&16.28/.8596&$<$11.72/.5899&$<$22.37/.6662&15.38/.2167&$<$21.22/.9614&23.31/.9433&$<$38.33/.9916&$<$8.33/.4929&$<$14.33/.4860\\
        H&31.02/.8997&17.92/.8977&20.36/.9364&21.37/.9030&33.17/.9481&23.68/.5453&26.75/.9864&34.22/.9702&43.16/.9942&12.45/.4771&17.39/.5260\\
        I&30.93/.8964&\underline{18.96/.9123}&21.94/.9527&20.64/.8939&32.87/.9447&24.17/.5713&28.84/.9908&{\bf 37.05}/{\bf .9768}&\underline{45.80/.9956}&12.46/.4589&16.99/.5241\\
         J&30.69/.8908&15.32/.8478&19.37/.9233&18.82/.8703&30.76/.9103&21.37/.4424&27.04/.9871&32.86/.9661&44.30/.9948&11.51/.4497&16.15/.5148\\
         K&28.57/.8369&13.76/.8121&18.33/.9069&16.36/.8131&22.42/.6713&{\bf 31.22}/{\bf .8385}&31.11/.9941&34.73/.9715&41.18/.9932&12.46/.4548&15.88/.5118\\
         L&31.26/.9040&16.37/.8693&17.59/.8918&18.64/.8667&22.37/.6663&27.91/.7189&21.75/.9725&24.51/.9457&38.64/.9918&11.74/.4468&15.70/.5028\\
         M&21.91/.6574&12.44/.7763&16.74/.8725&13.24/.6857&22.37/.6663&18.19/.3115&21.74/.9723&23.54/.9437&38.44/.9917&9.77/.3890&15.17/.4966\\
         N&29.73/.8686&12.97/.7891&19.03/.9193&15.91/.7979&29.22./8721&30.56/.8126&31.57/.9946&34.87/.9719&41.28/.9932&12.44/.4677&16.81/.5227\\
         O&21.58/.6518&7.94/.6033&16.89/.8761&12.38/.6371&22.37/.6662&15.47/.2203&21.51/.9632&23.95/.9446&38.55/.9917&8.26/.3593&15.23/.4957\\
        P&21.45/.6499&7.66/.5895&16.11/.8551&11.88/.5979&22.37/.6662&15.40/.2178&22.94/.9721&23.89/.9444&41.74/.9934&8.21/.3593&14.57/.4906\\
         \hdashline
    Q&{\bf 32.01}/{\bf .9147}&18.60/.9085&\underline{23.97/.9687}&{\bf 24.41}/{\bf .9262}&{\bf 33.49}/{\bf .9518}&\underline{30.56/.8166}&\underline{36.77/.9980}&34.42/.9702&44.27/.9948&\underline{16.54/.5351}&\underline{22.57/.5768}\\
         R&31.93/.9141&{\bf 19.13}/{\bf .9157}&{\bf 24.67}/{\bf .9724}&\underline{24.01/.9238}&\underline{33.30/.9502}&29.55/.7822&{\bf 37.62}/{\bf .9981}&\underline{36.95/.9765}&{\bf 46.09}/{\bf .9957}&{\bf 17.21}/{\bf .5577}&\textbf{23.22}/{\bf .5806}\\
         S&31.58/.9086&13.16/.7955&19.45/.9251&19.03/.8726&28.13/.8471&28.03/.7304&34.94/.9970&33.99/.9697&44.36/.9948&14.93/.5204&18.86/.5476\\
         T&31.32/.9055&12.82/.7877&19.24/.9214&16.69/.8235&29.83/.8867&27.01/.6878&36.13/.9976&32.51/.9651&43.82/.9947&10.97/.4303&15.88/.5077\\
        \hline
    \end{tabular}}
    \label{psnr-ssim}
\end{table*}

\begin{figure}[t]
\centerline{\includegraphics[width=1\columnwidth]{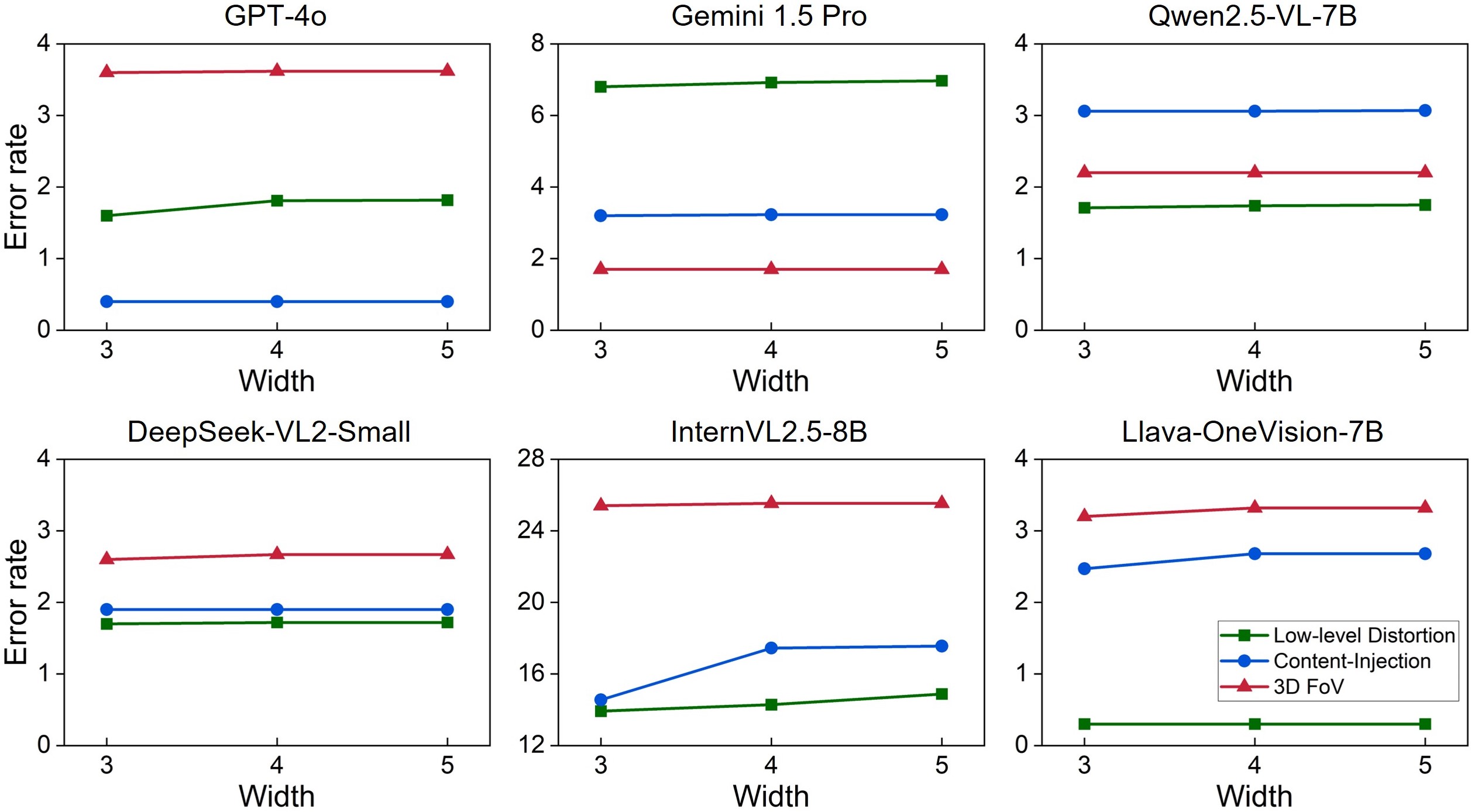}}
\caption{Comparative errors (\%) {\it w.r.t.} the width of the regularizer.}
\label{comparative-error}
\end{figure}

{\bf Implementation Details.}
Considering the trade-off between computational overhead and error rate, we set the width of the sliding window regularizer $w$ to $3$ (Fig. \ref{comparative-error}). For the NLI model, we adopt the BART model \cite{lewis2020bart} trained on the MultiNLI (MNLI) dataset \cite{williams2018broad} for the consideration of efficiency and effectiveness, which, theoretically, could be fulfilled by other appropriate large language models. 
Since the comparison order between image pairs is fixed, we first specify the reference positive and negative labels and then apply the NLI model to filter out contradictory analyses based on the probability of the label being true for positive and negative examples.
Moreover, apart from the proprietary LMMs that were deployed via API, all other experiments are conducted using a maximum of 8 Nvidia A800 80GB GPUs.

\begin{figure}[t]
\centerline{\includegraphics[width=1\columnwidth]{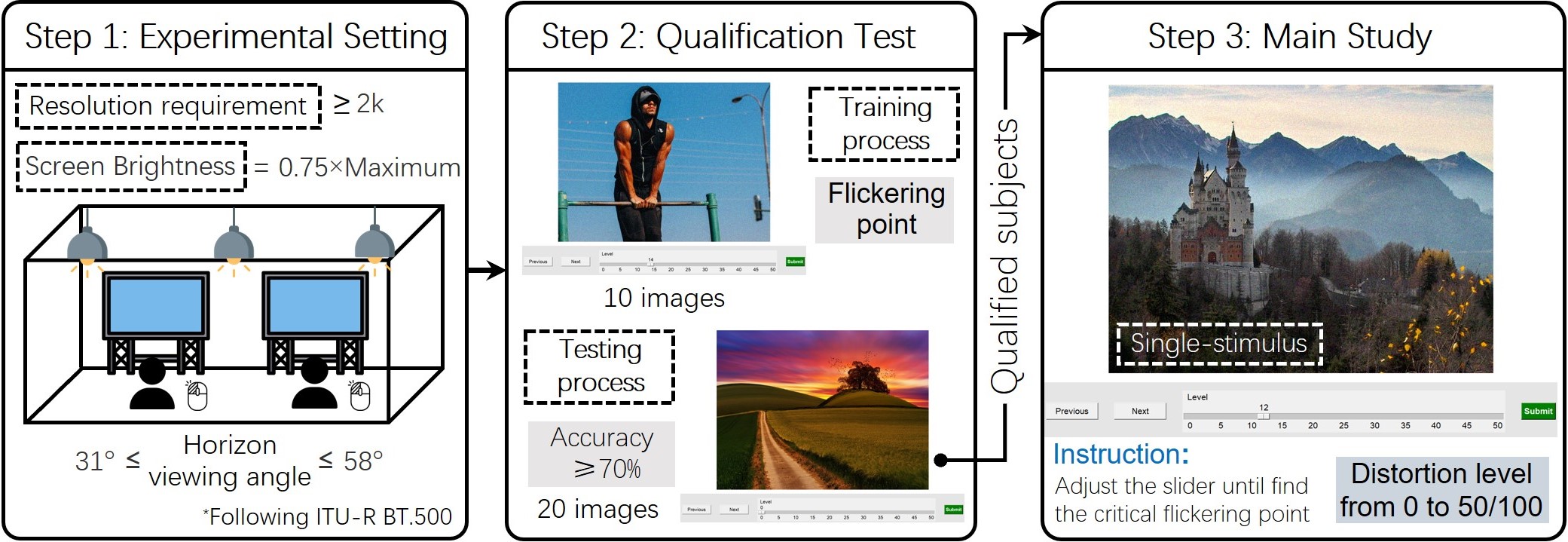}}
\caption{Workflow of the subjective JND assessment study. All experiments were conducted in an offline laboratory to ensure standardization.}
\label{workflow-human}
\end{figure}

{\bf Human-Level Performance.}
We conduct a subjective JND assessment study on the VPA-JND dataset to investigate the gap between humans and LMMs in visual acuity. 12 participants (8 male, 4 female, age=24.92$\pm$1.83) from campus, all with normal (corrected and no color blindness) eyesight, are recruited to independently answer each distortion group. Their performance, namely the smallest distortion level that humans can observe the image flicker, is collected following the procedures in \cite{lin2022large}.
Fig. \ref{workflow-human} illustrates the workflow of this study.
Specifically, we conduct the subjective studies in-lab to ensure that all participants have a clear and consistent understanding of the task. All images are displayed on two 27-inch screens with a resolution of 2560×1440.
Other settings, such as ambient brightness, lighting, and background, are configured according to the ITU-R BT.500 recommendation \cite{bt2002methodology}. 
After that, we provided subjects with 10 extra training images to familiarize them with the procedures ({\it e.g.}, how to use the interface). Participants were allowed into the main study only if they passed a quiz (consisting of another 20 images with different distortions), where we calculated the accuracy that at least 70\% slider positions should fall within 1.5 times the upper and lower limits of the ground truth ranges annotated by the expert group.

\subsection{Main Results}

{\bf Minimal Perception Granularity.} 
Tab. \ref{exp1} presents the $1$\textsuperscript{st} JND of LMMs evaluated on the proposed VPA-JND dataset.
First, human subjects achieve average minimal perceptible levels of 12.11 (except banding artifacts), 1.44, and 1.48 on low-level distortion, content-injection, and 3D FoV perspectives, lower than the best LMM by 26.2\%, 89.0\%, and 40.8\%, respectively. 
Notably, the performance gap between humans and the best LMM is much narrower on low-level distortion perception tasks due to the physiological characteristics of HVS, suggesting that LMM may have a relative strength in image quality-related low-level vision tasks.
Additionally, the leading LMM, Gemini 2.0 Flash, surpasses the other proprietary LMMs by a substantial margin and approaches human level in spatial awareness stimuli ({\it angle} and {\it distance}). 
We also observe that some open-source LMMs, such as InternVL2.5-78B/38B, Qwen2.5-VL-72B, and LLaVA-OneVision-72B, exhibit highly competitive performance to closed-source models, chasing or even leading Gemini 1.5 Pro by 13.6\% to 22.1\%. 

We notice that most small-scale LMMs, like those with 7B, 2B, or 0.5B parameters, struggle to respond differently to subtle signal variations (1\textsuperscript{st} JND is close to 50). 
Surprisingly, compared to its larger counterparts, Qwen2.5-VL-3B shows satisfactory performance on the {\it blur}, {\it JPEG}, {\it watermark}, and {\it distance} sets, which leads us to conduct further analyses of the model structure.
As for the model generation, Qwen2.5-VL exhibits better spatial perception capabilities than Qwen2-VL, since additional grounding data with absolute position coordinates is incorporated during the training process \cite{bai2025qwen25vl,wang2024qwen2}.
Tab. \ref{psnr-ssim} reports the PSNR and SSIM corresponding to the $1$\textsuperscript{st} JND for a more uniform comparison. Note that we omit the banding distortion since these two metrics are shown to have a low correlation with the banding severity \cite{chen2024band}.

\begin{figure}[t]
\centerline{\includegraphics[width=1\columnwidth]{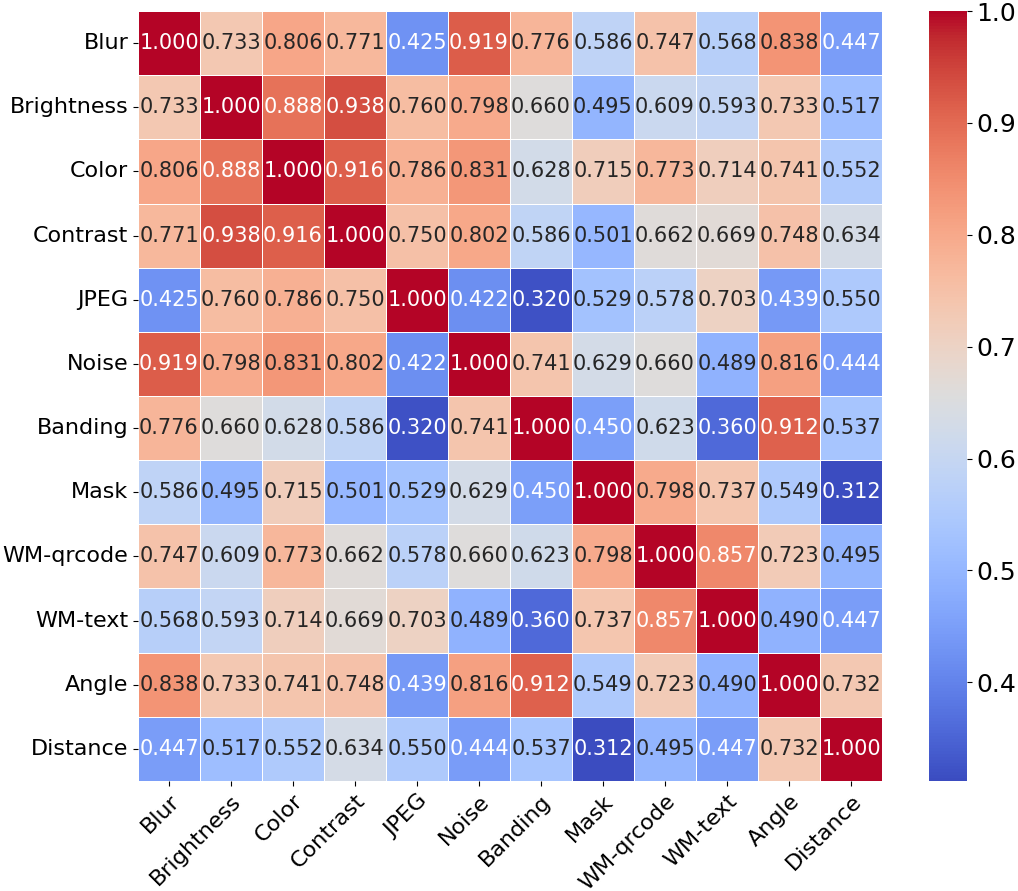}}
\caption{Pearson correlation between results across different dimensions.}
\label{correlation}
\end{figure}

\begin{figure*}[t]
\begin{center}
\includegraphics[width=1\linewidth]{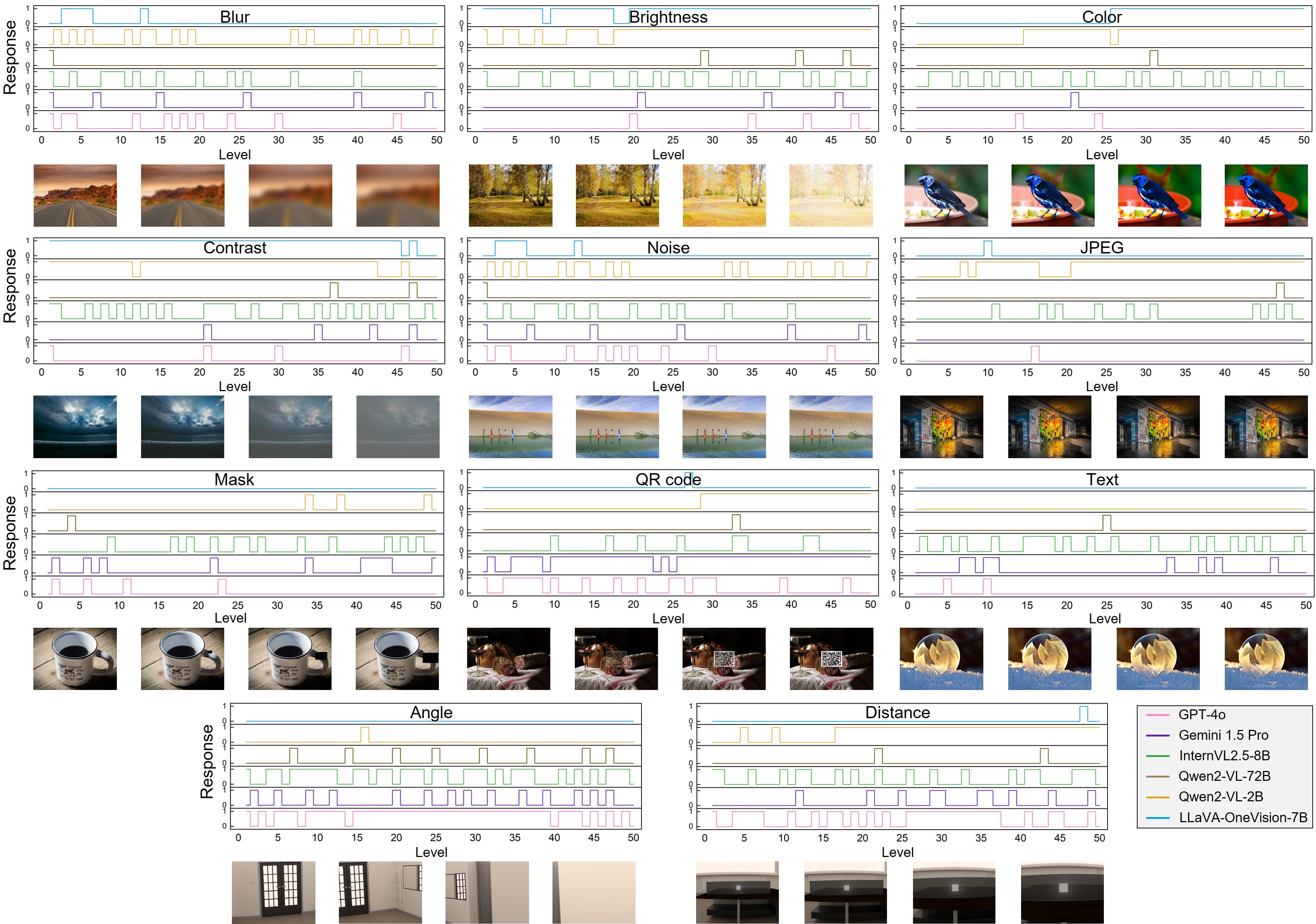}
\caption{LMM-JND curves of several randomly selected stimuli for six representative LMMs. The reference images and their distorted versions are displayed under the curve, where the leftmost one is the original reference image, and the rest are distorted variants with change level $=12,36,$ and $48$. The `0' and `1' on the vertical axis represent ``no response change'' and ``response changed'', respectively.}
\label{JND-curve}
\end{center}
\end{figure*}

\begin{figure}[t]
\centerline{\includegraphics[width=1\columnwidth]{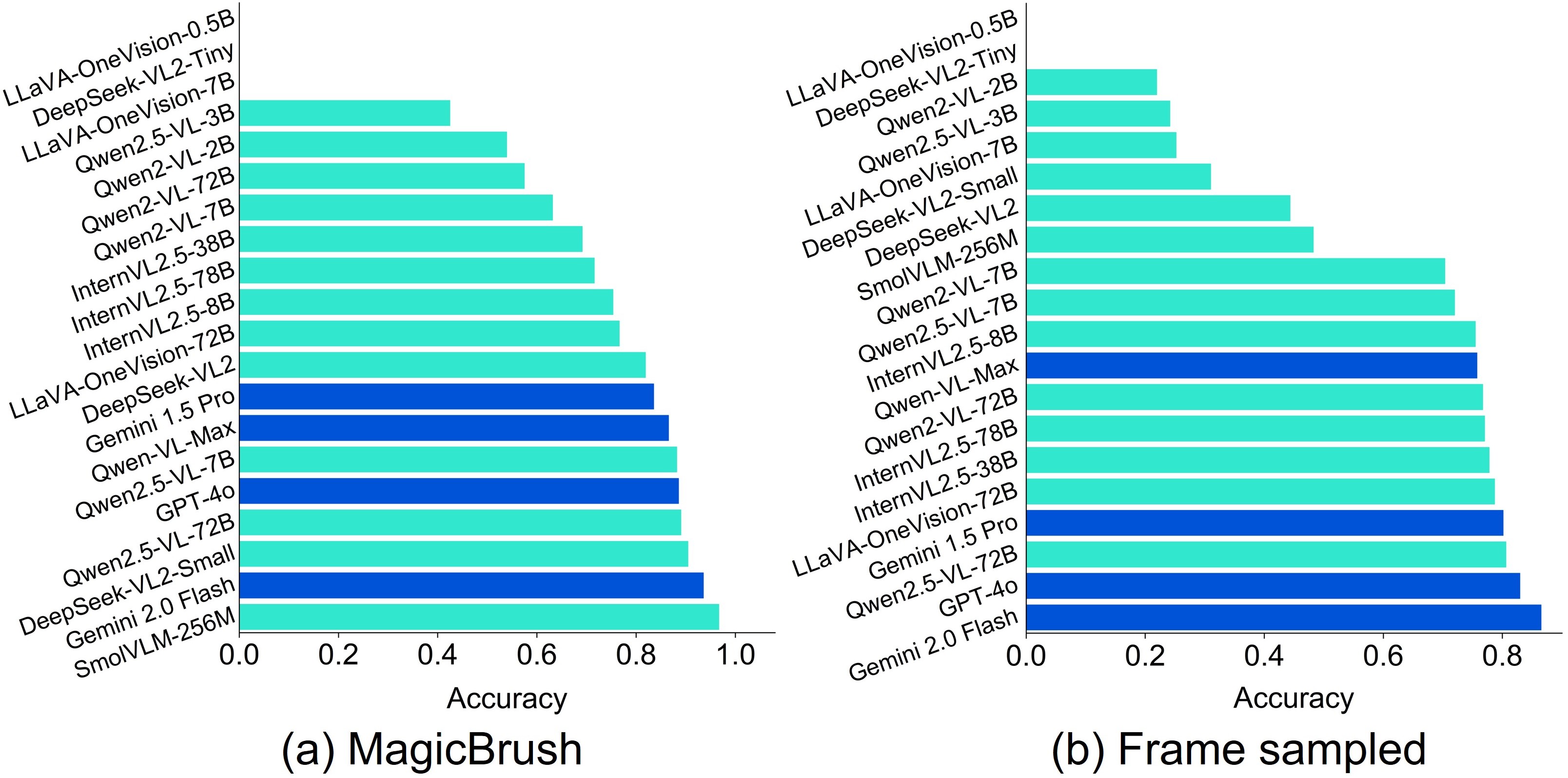}}
\caption{Benchmark results on two spot-the-difference test sets. Among them, Llava-OneVision-0.5B and DeepSeek-VL2-Tiny obtain less than 0.01\% accuracy in both test sets.}
\label{spot}
\end{figure}

{\bf $1$\textsuperscript{st} JND for different Stimuli.} From the perspective of different stimuli, the minimal perceptible distortion level for blur is significantly smaller than other low-level distortions, indicating a higher sensitivity of LMMs to the image structural changes. The higher PSNR and SSIM values in the blur column shown in Tab. \ref{psnr-ssim} also validate this conclusion.
In content-injection distortions, we find that LMMs are more effective at detecting QR code watermarks than flat masks and text-style watermarks, possibly due to their unique shape features.
For spatial 3D FoV stimuli, LMMs demonstrate greater proficiency in detecting angular variations compared to distance changes that involve more depth information processing in the visual scene. 

We further investigate the correlations between different stimulus types as shown in Fig. \ref{correlation}.  It can be observed that the mask distortion and spatial distance perception possess significantly low correlations with other stimuli, highlighting the content uniqueness of their stimuli. Moreover, the low-level distortions manifest strong inter-correlation yet are distinctly independent of other distortion types, which in turn validates the construction of our VPA-JND dataset. Unexpectedly, banding shows a relatively lower correlation with the other low-level distortions (avg. Pearson's $r\approx0.619$). We speculate that this is due to its quantization rule being based on subjective experiments, which is fundamentally different from mathematical quantization.

{\bf Redundancy Zone Observation.} We select six representative LMMs and plot their LMM-JND curves, namely the $n$\textsuperscript{th} JND, of different stimuli in Fig. \ref{JND-curve} to give a first impression of what LMM-JND curves look like. 
The width of the trough between two high-level peaks represents the perceptual redundancy interval of LMM. 
We observe distinct perceptual redundancy intervals for all types of stimulus in two proprietary LMMs and Qwen2-VL-72B. 
In comparison, InternVL2.5-8B displays more frequent fluctuations in response, which is consistent with the findings in pilot experiments (Tab. \ref{Sec3::exp2}), demonstrating its fine-grained perception ability (same in Tab. \ref{exp1})
Interestingly, there can be dense response variations even in low or high distortion levels, indicating that the magnitude of signal change and LMM-JND do not have a monotonic correlation. 
In other words, the same degree of stimulus variation under high distortion levels may lead to larger LMM-JNDs ({\it e.g.}, GPT-4o in {\it blur} and {\it noise} scenarios), implying that the lower signal quality could result in more consistent model perceptions.
Furthermore, it reveals the possibility of achieving controllable robustness or security monitoring in LMMs by utilizing such perceptual blind zone.

\begin{figure*}[t]
\centerline{\includegraphics[width=1\linewidth]{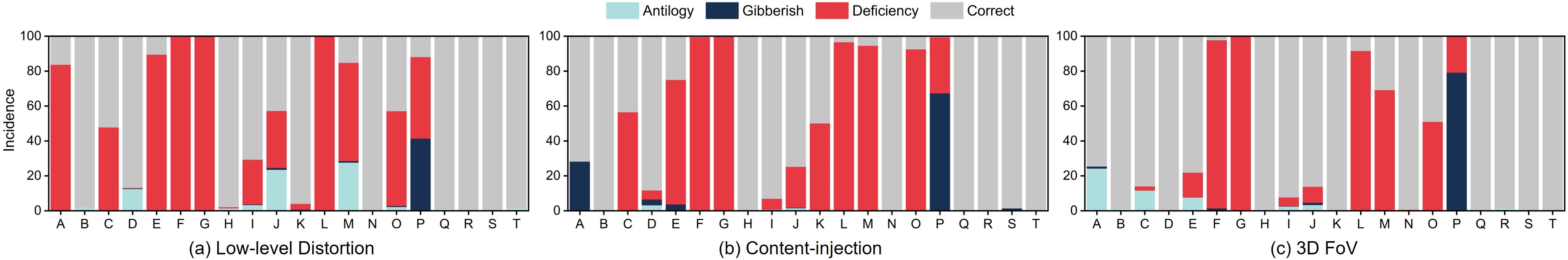}}
\caption{The incidence (\%) of problematic responses during LMM-JND determination. The values are split by antilogy, gibberish, deficiency, and correct results.  }
\label{error-analysis}

\end{figure*}

\begin{figure*}[t]
\centerline{\includegraphics[width=1\linewidth]{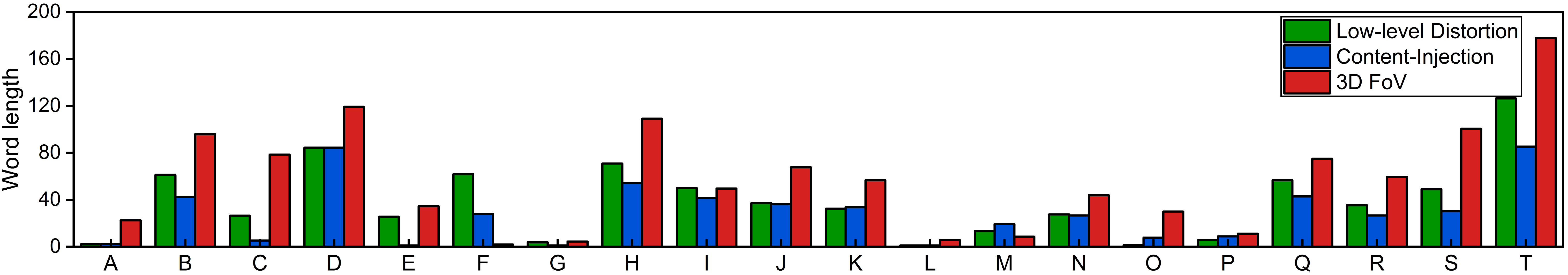}}
\caption{The average word length in the responses of 20 LMMs evaluated in terms of low-level distortion, content-injection, and 3D FoV perspectives.}
\label{word_length}
\end{figure*}

\begin{figure*}[t]
\begin{center}
\includegraphics[width=1\linewidth]{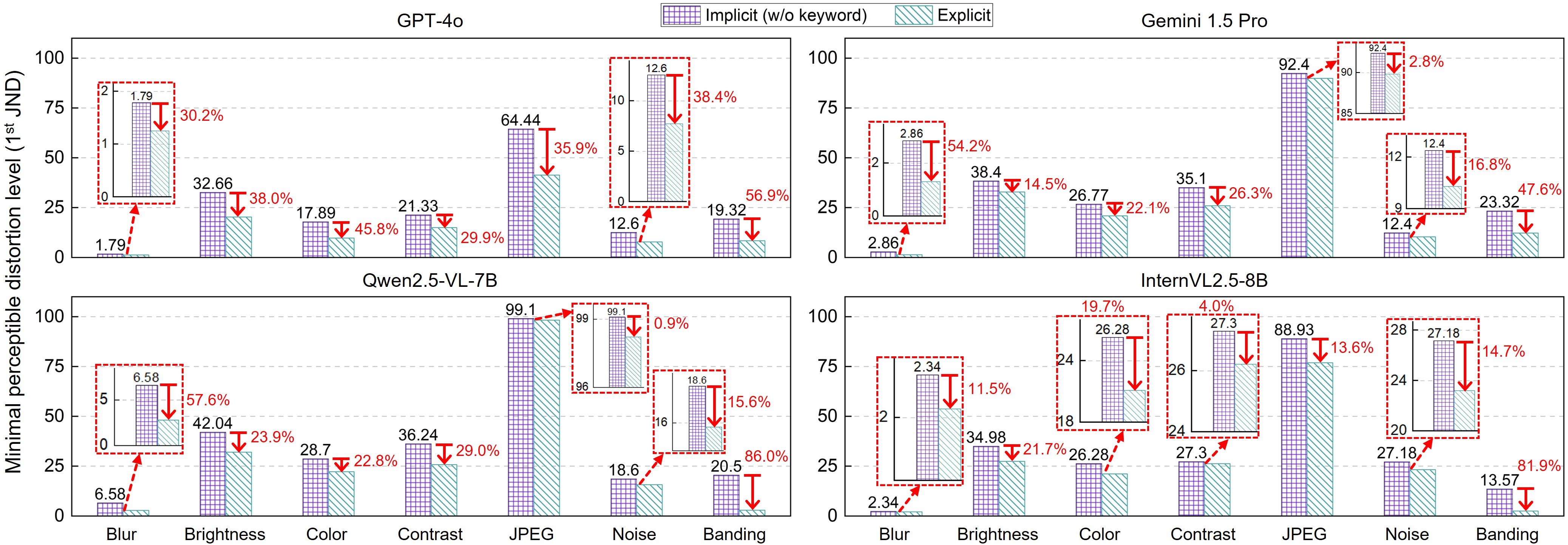}
\caption{Ablation experiments on the enhancement of perceptual granularity after applying explicit prompting for low-level stimuli.}
\label{prompting}
\end{center}
\end{figure*}

{\bf Spot the Difference.} Apart from formulaic constructed stimuli, we also evaluate the intentional scenarios for LMMs, where contents are more deliberately hidden, shown in Fig. \ref{spot}.
In general, the performance of proprietary LMMs remains at the forefront, and open-source models such as Qwen2.5-VL-72B are also demonstrating impressive capabilities.
Surprisingly, the smallest SmolVLM-256M performs the best in the MagicBrush set, surpassing models that are several orders of magnitude larger than it. We hypothesize that the adopted large image patch size of $512\times 512$ in the vision encoder allows the information to be encoded more efficiently, even with only 93M parameters (Tab. \ref{tab:arch}).
However, in the frame-sampled set where inter-image differences are more subtle and image quality is lower, SmolVLM-256M only achieves about 50\% accuracy, and LMMs with larger vision backbones remain robust.
Meanwhile, we find the performance disparity between large- and small-scale models is more significant on the frame-sampled set compared to MagicBrush, indicating that fine-grained perception is more influenced by scaling laws than general visual semantic understanding.
Besides, LLaVA-OneVision-0.5B and DeepSeek-VL2-Tiny score 0\% on both sets, together with the former experiments, showing their severe inferiority in detailed perception.
To sum up, the outcomes on these low-complexity spot-the-difference tasks suggest that, irrespective of model size or training data, current LMMs struggle with subtle visual details (avg. acc. $<0.8$), which may not be a knowledge base limitation.

\subsection{Analysis of LMM-JND Acquisition Process}
{\bf Response Error Analysis.} We further report the occurrence rate of response error during the JND determination process in Fig. \ref{error-analysis}. Their qualitative examples are pre-given in Fig. \ref{example-hallu}. First, we observe a large proportion on deficiency outputs for all three perspectives, especially in DeepSeek-VL2 family, Qwen2-VL-\{2B, 7B\}, and Llava-OneVision-\{0.5B, 7B\}, where over 80\% of responses are detected as invalid, resulting in higher 1\textsuperscript{st} JND while reflecting their poor detail perception capability.
Second, antilogical responses are more likely to occur in InternVL2.5 series and small models (Param.$\leq$3B), highlighting their instability. Similarly, SmolVLM-256M and LLaVA-OneVision-0.5B with extremely small language backbones are prone to outputting gibberish.
Third, larger open-source models (Param.$\geq$72B) and proprietary models seldom produce errors in their output, demonstrating their powerful language backbones. 
Fourth, the problematic responses on 3D FoV set are less than on low-level distortion and content-injection, emphasizing the impact of the task complexity that LMMs may instead become less reliable on simpler tasks.
As seen from Fig. \ref{comparative-error}, the response length of large open-source models and proprietary models is significantly longer than small models, suggesting a greater tendency to generate complete and analytical outputs.
This trend is also evident in the timing of model releases.

{\bf Comparative Error vs. the Width of Regularizer.} In Fig. \ref{comparative-error}, we discuss the width of the regularizer during the LMM-JND determination process, {\it i.e.} the issue of monotonicity in perception identified through additional scrutiny within the regularizer, with comparative errors reported. Six representative models are selected for illustration. We find that the comparative error shows almost no variation under different window widths, indicating an acceptable value to balance the precision of LMM-JND points and computational overhead. From a numerical perspective, the error rate of InternVL2.5-8B is substantially higher, which corroborates the instability observed in the above experiments.

{\bf Necessity of Explicit Prompting.} In Fig. \ref{prompting}, we compare the minimal perceptible level of four LMMs under implicit and explicit prompting (Fig. \ref{pipeline}), {\it i.e.}, whether the distortion type {\tt <dist.>} is mentioned in the questions. It is evident that all LMMs obtain perceptual granularity improvement through explicit prompting for the target stimulus, illustrating that LMMs generally focus more on semantic information than minor low-level distortions.
Among all low-level distortions, the performance gains in blur and banding sets are the most pronounced, where the $1$\textsuperscript{st} JND decreases on average 38.4\% and 68.1\%, respectively. 
This further highlights the necessity of this design in LMM-JND acquisition.

\begin{figure}[t]
\centerline{\includegraphics[width=1\columnwidth]{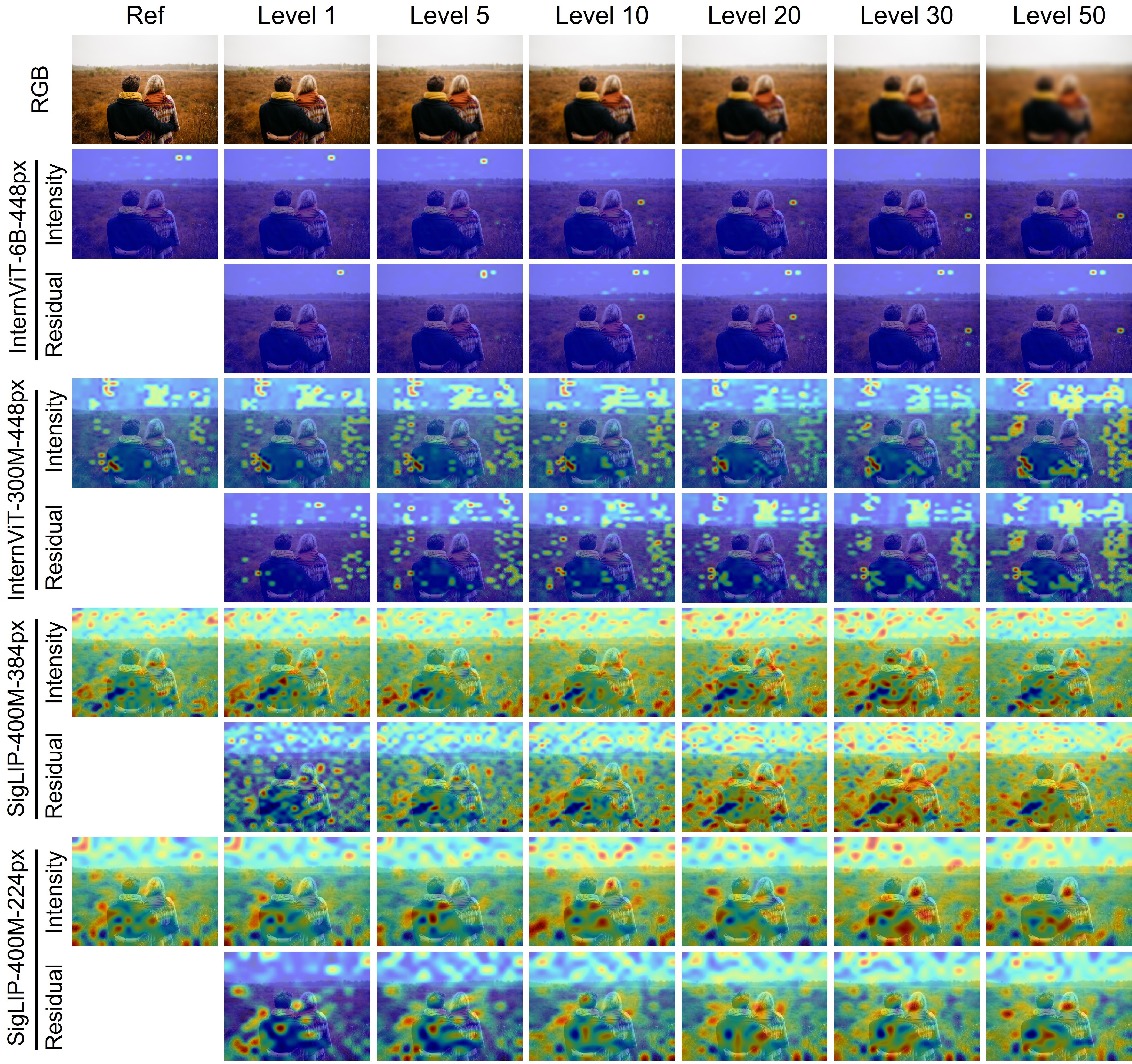}}
\caption{Visualization of the feature intensity distribution of different stimuli ({\it blur}) encoded by different visual backbones, as well as their feature residuals.}
\label{vision-encoder}
\end{figure}

\subsection{In-depth Analysis}

\begin{figure}[t]
\centerline{\includegraphics[width=1\columnwidth]{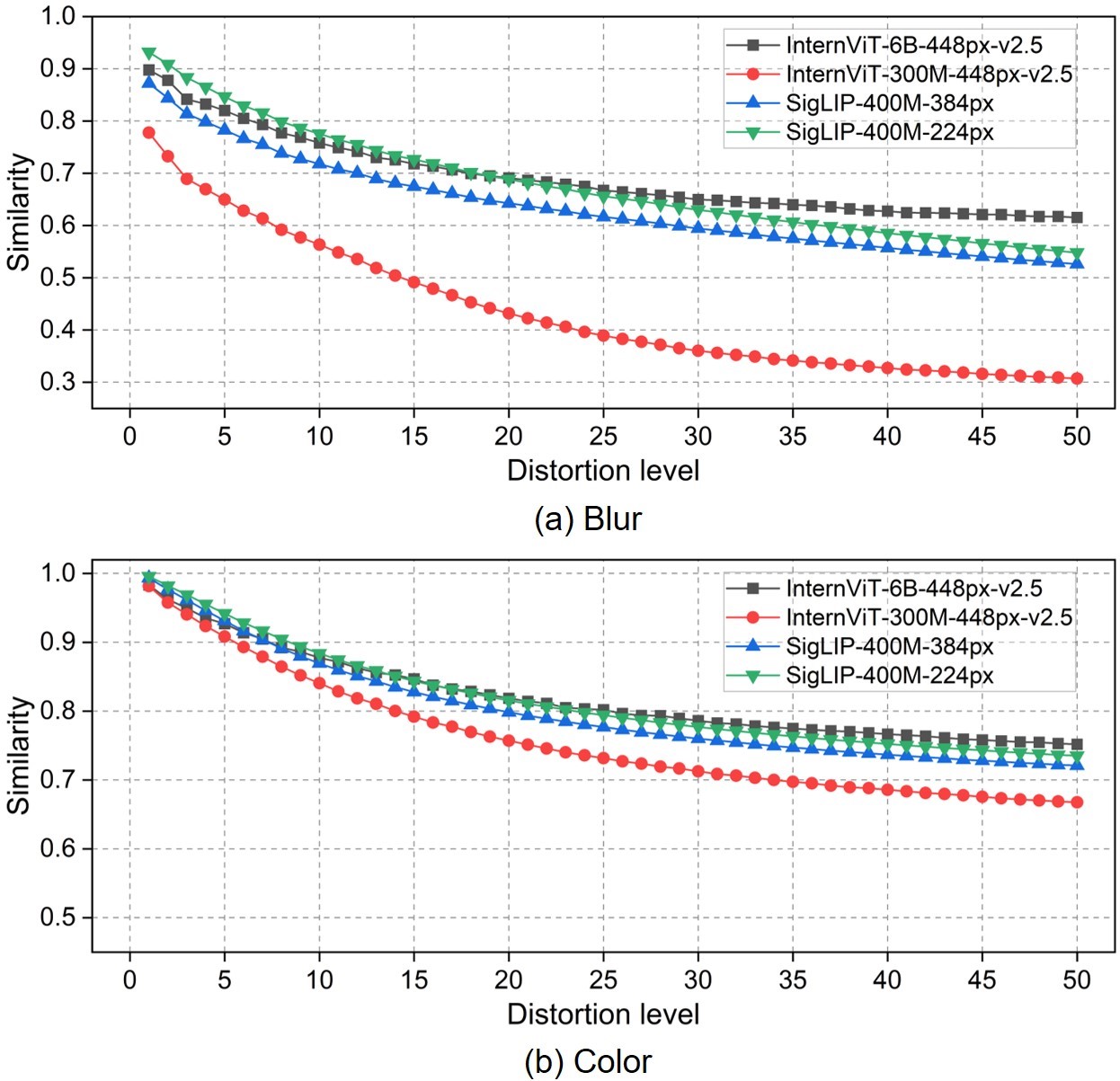}}
\caption{The average cosine similarity between the visual embeddings of 1,008 reference images and their counterparts with (a) blur and (b) color distortions.}
\label{cosine-similarity}
\end{figure}

{\bf Impact of Visual Encoder.} 
In Fig. \ref{vision-encoder}, we decouple the visual encoder from LMMs and visualize the feature intensity distribution for the reference image and blur-distorted stimuli. Four mainstream vision backbones are studied, including InternViT-6B-v2.5-448px, InternViT-300M-v2.5-448px, SigLip-400M-384px, and SigLip-400M-224px.
Based on this, we gain three key observations. First, as the distortion level increases, the activation regions of the model deviate gradually from those of the original image (see the residual feature maps). 
Second, a larger model size ({\it e.g.}, 6B vs. 300M or 400M) leads to a highly focused and sparse distribution of activation maps, whereas smaller models exhibit a more diffuse and dense distribution, possibly due to the perceived coarser feature granularity.
Third, as shown in the sizes of the local activation area, the patch size directly affects the perception granularity ({\it e.g.}, 448px vs. 384px or 224px). 
Besides, Fig. \ref{cosine-similarity} quantitatively validates the perceptual deficiencies of these visual encoders towards low-level distortions, where distinct feature differences are only observed at high distortion levels. We also notice that all curves exhibit exponential decay, and InternViT-300M-v2.5-448px is the most sensitive to distortion changes, aligning with the residual maps in Fig. \ref{vision-encoder}.

\begin{table}[t]
    \centering
    \renewcommand\arraystretch{1.1}
    \caption{The Correlation between the Language Backbone Size and $1$\textsuperscript{st} JND for Different Perspectives in VPA-JND Dataset}
    \resizebox{1\linewidth}{!}{\begin{tabular}{c|c|ccc}
    \hline                 
        {\textbf{Model (series)}}  &$\mathrm{Param}_L$& Low-level&Content-Injection&3D FoV \\ \hline
         Qwen2.5-VL&Qwen2.5-72B/7B/3B&0.9971&0.9328&0.1233\\
        DeepSeek-VL2&DeepSeekMoE-27B/16B/3B&0.8656&0.8428&0.5271\\
        Qwen2-VL&Qwen2-72B/7B/1.5B&0.8816&0.9999&0.6541\\
         LLaVA-OV&Qwen2-72B/7B/0.5B&0.9973&0.9394&0.9990\\
         \hline
    \end{tabular}}
    \label{PLCC-Lsize}
\end{table}

\begin{figure*}[t]
\setlength{\abovecaptionskip}{0.cm}
\centerline{\includegraphics[width=1\linewidth]{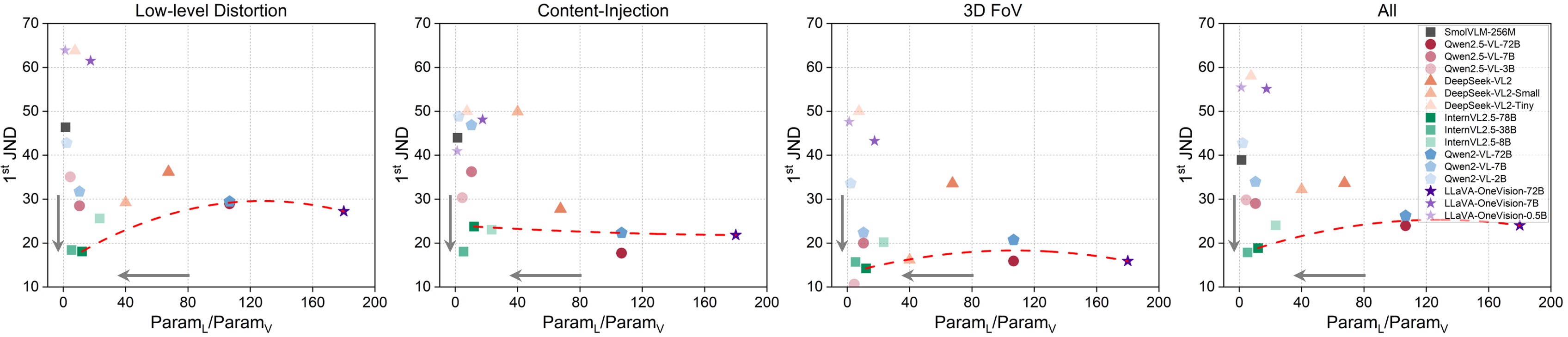}}
\caption{$1$\textsuperscript{st} JND versus the parameter ratio of the language backbone ($\mathrm{Param}_L$) to the visual backbone ($\mathrm{Param}_V$). The red dotted curves show the fitted optimization direction of the perception granularity for large models (72B/78B). }
\label{scatter-ratio}
\end{figure*}

{\bf Language Backbone vs. $1$\textsuperscript{st} JND.}
Beyond vision components, we examine the perceptual granularity of LMMs from the language backbone. Specifically, four LMM groups are selected, {\it i.e.}, Qwen2.5-VL, DeepSeek-VL2, Qwen2-VL, and LLaVA-OneVision, each equipped with identical visual encoders. 
As listed in Tab. \ref{PLCC-Lsize}, the performance is essentially proportional to the model scale, except for Qwen2.5-VL in the 3D FoV set.
Intuitively, we further wonder whether there exists a {\it golden ratio} between vision and language backbone size. 
As shown in Fig. \ref{scatter-ratio}, LMMs with stronger detail perception capabilities generally have a lower $\mathrm{Param}_L/\mathrm{Param}_V$, providing critical insights for architectural refinement of future LMMs.
Based on the previous experiments, we can conclude that given a sufficiently large language backbone to ensure low response error, a larger visual backbone leads to finer-grained visual perception.

\begin{table*}[t]
    \centering
    \renewcommand\arraystretch{1.45}
    \caption{The Imperceptible Proportion after Injecting Different Levels of Other Distortions below the Respective LMM-JND for Qwen2.5-VL-72B (upper) and Gemini 1.5 Pro (lower)}
    \resizebox{1\linewidth}{!}{\begin{tabular}{c|ccccc|ccccc}
    \hline                 
        {\textbf{Source\textbackslash Injected Level}}&$\frac{1}{5} \left\lfloor \mathcal{MRV}_{col.} \right\rfloor$&$\frac{2}{5} \left\lfloor \mathcal{MRV}_{col.} \right\rfloor$ &$\frac{3}{5} \left\lfloor \mathcal{MRV}_{col.} \right\rfloor$&$\frac{4}{5} \left\lfloor \mathcal{MRV}_{col.} \right\rfloor$&$\left\lfloor \mathcal{MRV}_{col.} \right\rfloor$&$\frac{1}{5} \left\lfloor \mathcal{MRV}_{contr.} \right\rfloor$&$\frac{2}{5} \left\lfloor \mathcal{MRV}_{contr.} \right\rfloor$&$\frac{3}{5} \left\lfloor \mathcal{MRV}_{contr.} \right\rfloor$&$\frac{4}{5} \left\lfloor \mathcal{MRV}_{contr.} \right\rfloor$&$\left\lfloor \mathcal{MRV}_{contr.} \right\rfloor$ \\ \hline
        $\frac{1}{5} \left\lfloor \mathcal{MRV}_{bri.} \right\rfloor$&100\%&100\%&100\%&100\%&94.17\%&100\%&100\%&100\%&100\%&92.23\%\\
        $\frac{2}{5} \left\lfloor \mathcal{MRV}_{bri.} \right\rfloor$&100\%&99.03\%&99.03\%&98.06\%&87.38\%&100\%&100\%&100\%&100\%&85.44\%\\
        $\frac{3}{5} \left\lfloor \mathcal{MRV}_{bri.} \right\rfloor$&99.03\%&97.09\%&94.17\%&82.52\%&77.67\%&99.03\%&99.03\%&100\%&99.03\%&73.79\%\\
        $\frac{4}{5} \left\lfloor \mathcal{MRV}_{bri.} \right\rfloor$&94.17\%&91.26\%&82.52\%&74.76\%&59.22\%&96.12\%&96.12\%&98.06\%&95.15\%&55.34\%\\
         $\left\lfloor \mathcal{MRV}_{bri.} \right\rfloor$&82.52\%&70.87\%&66.99\%&55.34\%&45.63\%&83.50\%&82.52\%&74.76\%&56.31\%&21.36\%\\
         \hline
         \hline                 
        {\textbf{Source\textbackslash Injected Level}}&$\frac{1}{5} \left\lfloor \mathcal{MRV}_{col.} \right\rfloor$&$\frac{2}{5} \left \lfloor \mathcal{MRV}_{col.} \right\rfloor$ &$\frac{3}{5} \left\lfloor \mathcal{MRV}_{col.} \right\rfloor$&$\frac{4}{5} \left\lfloor \mathcal{MRV}_{col.} \right\rfloor$&$\left\lfloor \mathcal{MRV}_{col.} \right\rfloor$&$\frac{1}{5} \left\lfloor \mathcal{MRV}_{contr.} \right\rfloor$&$\frac{2}{5} \left\lfloor \mathcal{MRV}_{contr.} \right\rfloor$&$\frac{3}{5} \left\lfloor \mathcal{MRV}_{contr.} \right\rfloor$&$\frac{4}{5} \left\lfloor \mathcal{MRV}_{contr.} \right\rfloor$&$\left\lfloor \mathcal{MRV}_{contr.} \right\rfloor$ \\ \hline
        $\frac{1}{5} \left\lfloor \mathcal{MRV}_{bri.} \right\rfloor$&98.06\%&97.09\%&85.44\%&78.64\%&56.31\%&42.72\%&40.78\%&44.66\%&42.72\%&34.95\%\\
        $\frac{2}{5} \left\lfloor \mathcal{MRV}_{bri.} \right\rfloor$&92.23\%&84.47\%&84.47\%&55.34\%&43.69\%&44.60\%&43.69\%&39.81\%&40.78\%&33.98\%\\
        $\frac{3}{5} \left\lfloor \mathcal{MRV}_{bri.} \right\rfloor$&83.50\%&79.61\%&71.84\%&54.37\%&39.81\%&38.83\%&38.83\%&31.07\%&34.95\%&26.21\%\\
        $\frac{4}{5} \left\lfloor \mathcal{MRV}_{bri.} \right\rfloor$&68.93\%&55.34\%&52.43\%&39.81\%&29.13\%&30.10\%&28.16\%&24.27\%&21.07\%&15.53\%\\
         $\left\lfloor \mathcal{MRV}_{bri.} \right\rfloor$&43.69\%&39.81\%&33.98\%&33.01\%&21.36\%&18.45\%&18.45\%&16.50\%&16.50\%&9.71\%\\
         \hline
    \end{tabular}}
    \label{homo1}
\end{table*}

\begin{table*}[t]
    \centering
    \renewcommand\arraystretch{1.45}
    \caption{The Imperceptible Proportion after Injecting Different Levels of Other Distortions between LMM-JND Points for Qwen2.5-VL-72B (upper) and Gemini 1.5 Pro (lower). $N$ Denotes the Size of Test Sets}
    \resizebox{1\linewidth}{!}{\begin{tabular}{c|ccccc|ccccc}
    \hline                 
        {\textbf{Ref (Bri.)\textbackslash Injected}}  &$\frac{1}{5} \left\lfloor \mathcal{MRV}^{n}_{col.} \right\rfloor$& $\frac{2}{5} \left\lfloor \mathcal{MRV}^{n}_{col.} \right\rfloor$&$\frac{3}{5} \left\lfloor \mathcal{MRV}^{n}_{col.} \right\rfloor$&$\frac{4}{5} \left\lfloor \mathcal{MRV}^{n}_{col.} \right\rfloor$&$\left\lfloor\mathcal{MRV}^{n}_{col.} \right\rfloor$&$\frac{1}{5} \left\lfloor \mathcal{MRV}^{n}_{contr.} \right\rfloor$& $\frac{2}{5} \left\lfloor \mathcal{MRV}^{n}_{contr.} \right\rfloor$&$\frac{3}{5} \left\lfloor \mathcal{MRV}^{n}_{contr.} \right\rfloor$&$\frac{4}{5} \left\lfloor \mathcal{MRV}^{n}_{contr.} \right\rfloor$&$\left\lfloor\mathcal{MRV}^{n}_{contr.} \right\rfloor$ \\ \hline
         2\textsuperscript{nd} LMM-JND ($N=489$)&100\%&100\%&100\%&100\%&100\%&100\%&100\%&100\%&100\%&99.18\%\\
        3\textsuperscript{rd} LMM-JND ($N=127$)&100\%&100\%&100\%&100\%&100\%&100\%&100\%&100\%&99.21\%&92.13\%\\
         \hline
         \hline 
         {\textbf{Ref (Bri.)\textbackslash Injected}}  &$\frac{1}{5} \left\lfloor \mathcal{MRV}^{n}_{col.} \right\rfloor$& $\frac{2}{5} \left\lfloor \mathcal{MRV}^{n}_{col.} \right\rfloor$&$\frac{3}{5} \left\lfloor \mathcal{MRV}^{n}_{col.} \right\rfloor$&$\frac{4}{5} \left\lfloor \mathcal{MRV}^{n}_{col.} \right\rfloor$&$\left\lfloor\mathcal{MRV}^{n}_{col.} \right\rfloor$&$\frac{1}{5} \left\lfloor \mathcal{MRV}^{n}_{contr.} \right\rfloor$& $\frac{2}{5} \left\lfloor \mathcal{MRV}^{n}_{contr.} \right\rfloor$&$\frac{3}{5} \left\lfloor \mathcal{MRV}^{n}_{contr.} \right\rfloor$&$\frac{4}{5} \left\lfloor \mathcal{MRV}^{n}_{contr.} \right\rfloor$&$\left\lfloor\mathcal{MRV}^{n}_{contr.} \right\rfloor$ \\ \hline
         2\textsuperscript{nd} LMM-JND ($N=697$)&100\%&93.97\%&82.93\%&78.05\%&71.31\%&99.57\%&94.26\%&83.36\%&69.01\%&63.85\%\\
        3\textsuperscript{rd} LMM-JND ($N=633$)&100\%&98.10\%&94.94\%&85.47\%&79.15\%&100\%&95.89\%&82.62\%&69.35\%&65.09\%\\
         \hline
    \end{tabular}}
    \label{homo2}
\end{table*}

\begin{table*}[t]
    \centering
    \renewcommand\arraystretch{1.05}
    \caption{The Response Change Ratio (\%) after JPEG Compression and the Average Saved Bits Per Pixel (bpp) Separated by Slash}
    \resizebox{1\linewidth}{!}{\begin{tabular}{l|cccccccccccccc}
    \hline                 
        {\textbf{Model (Compre. Level)}}&Existence& Count&Position& Color&OCR& Poster&Celebrity&Scene&Landmark&Artwork&Comm.&Num.&Text.&Code. \\ \hline
        Qwen2.5-VL-7B (98)&6.7/1.242&21.7/1.329&21.7/1.187&23.3/1.313&7.5/0.686&6.5/1.277&32.6/2.285&11.0/1.163&16.5/2.066&22.5/1.064&9.3/8.155&12.5/10.118&0/0.360&0/1.037\\
        InternVL2.5-78B (46)&3.3/0.172&10.0/0.194&25.0/0.174&6.7/0.189&5.0/0.088&8.8/0.528&14.7/1.752&6.8/0.170&14.8/1.332&13.3/0.114&15.0/7.669&25.0/9.543&7.5/-0.581&2.5/0.675\\
        Llava-OneVision-72B (85)&0/0.996&0/1.052&2.3/0.939&0/1.043&0/0.576&0/1.026&7.2/2.111&5.3/0.910&0/1.814&11.4/0.837&0/7.981&0/9.975&0/0.008&0/0.902\\
        GPT-4o (41)&0/0.148&0/0.166&0/0.147&0/0.161&0/0.073&0/0.464&0/1.700&0/0.144&0/1.209&0/0.063&0/7.626&0/9.477&0/-0.656&0/0.645\\
        Gemini 2.0 Flash (44)&0/0.159&0/0.179&0/0.160&0/0.174&0/0.080&0/0.505&0/1.730&0/0.157&0/1.305&0/0.076&0/7.650&0/9.515&0/-0.612&0/0.662\\
         \hline
    \end{tabular}}
    \label{JPEG}
\end{table*}

\begin{figure*}[t]
\setlength{\abovecaptionskip}{0.cm}
\centerline{\includegraphics[width=1\linewidth]{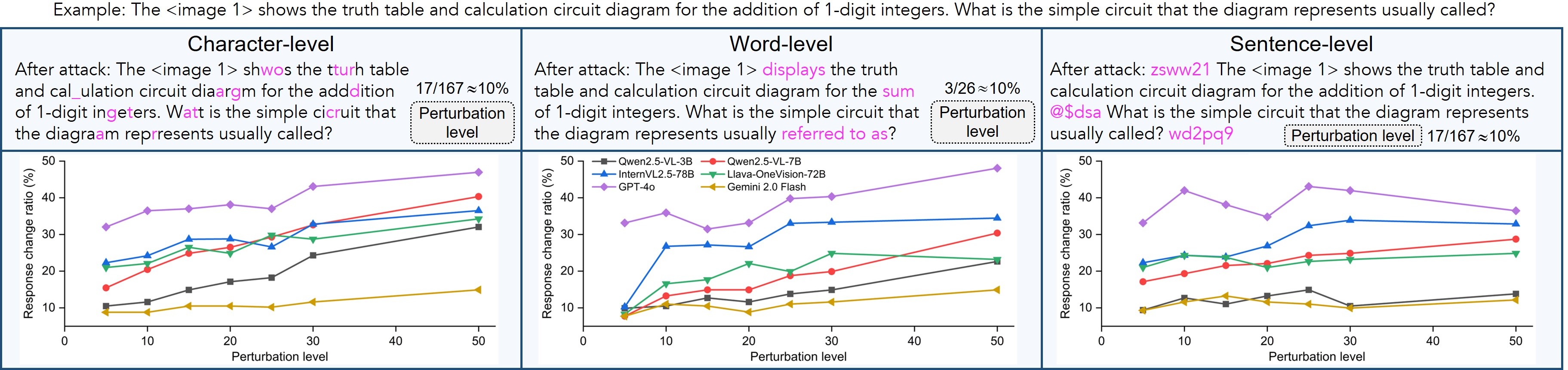}}
\caption{Illustration of different textual attacks and the response change ratio {\it w.r.t.} the perturbation level. The image placeholders are excluded in counting the characters and words.}
\label{text-JND}
\end{figure*}

\subsection{Homogeneous Property Test}
One of the key properties of the HVS-JND is its homogeneity, which refers to its inability to perceive signal changes below the HVS-JND threshold that is widely utilized to guide the efficient compression and encoding in multimedia systems \cite{jayant1993signal,lin2021progress}. In this section, we investigate whether LMMs have such a homogeneous property in cross-distortion scenarios. One proprietary model (Gemini 1.5 Pro) and one open-source model (Qwen2.5-VL-72B) are selected for evaluation.

As demonstrated by the experimental results presented in Tab. \ref{exp1} and Fig. \ref{JND-curve}, the LMM-JND points exhibit a dependency on both distortion type and content, and they vary significantly across different model structures.
To this end, we design two schemes evaluating on three types of low-level distortions ({\it i.e.} brightness, color, and contrast) as follows.
\begin{itemize}
    \item {\it Content-oriented LMM-JND Contamination:} We first filter out those images with all 1\textsuperscript{st} LMM-JNDs higher than the average value of the respective dimension from 1,008 reference images, that is, the perceptually more challenging samples, and then apply composite distortion to generate the stimuli. 
    Concretely, we create 5 new below-LMM-JND images, which have been exacerbated by $\frac{1}{5}$, $\frac{2}{5}$, $\frac{3}{5}$, $\frac{4}{5}$, and one times of $\left\lfloor \mathcal{MRV} \right\rfloor$ levels for each distortion, respectively, as the injection recipients or sources, where $\left\lfloor \cdot\right\rfloor$ denotes the floor operation. 
    Tab. \ref{homo1} reports the remaining imperceptible proportion $\tau$ for different injection combinations. 
    We observe that Qwen2.5-VL-72B cannot perceive any changes ($\tau=100\%$) under the settings that combined distortions do not exceed the LMM-JND. Besides, Gemini 2.0 Flash is more sensitive to combined distortions ($\tau<85\%$ when confronting $\frac{2}{5} \left\lfloor \mathcal{MRV}_{bri.} \right\rfloor$ + $\frac{2}{5} \left\lfloor \mathcal{MRV}_{contr.} \right\rfloor$), especially to brightness-contrast combinations than to brightness-color combinations.
    \item {\it Contamination between LMM-JND Points:} Here, we choose images that have multiple LMM-JND points within the distortion range of this study to form the test sets, and then apply distortions with $\frac{1}{5}$, $\frac{2}{5}$, $\frac{3}{5}$, $\frac{4}{5}$, and one times of the corresponding $\left\lfloor \mathcal{MRV}^{n} \right\rfloor$ levels to generate stimuli, where $n$ is the order of LMM-JND. As shown in Tab. \ref{homo2}, the homogeneous property between LMM-JND points are prominently exhibited on both Qwen2.5-VL-72B and Gemini 2.0 Flash, showing a certain degree of tolerance between distortions. 
\end{itemize}
Through these experiments, we demonstrate that the generated LMM-JND possesses similar homogeneity characteristics to HVS-JND, {\it i.e.}, distortions below the LMM-JND threshold can be tolerated by the LMM. However, compared to the  commonality observed within human populations, the performance discrepancies among different LMMs highlight their individual variability, making their uniqueness more pronounced.

\subsection{LMM-JND-Guided Image Compression}
Besides pixel-level evaluation, we further probe LMM-JND in content understanding tasks, which present semantic-level and practical challenges in open-world applications, such as large-scale visual question answering (VQA) benchmarks and online VQA platforms, that require runtime, throughput, and storage space considerations.
Since LMM-JND implies the visibility limitation of LMMs, we conduct experiments to validate whether it can improve compression efficiency while not affecting normal VQA. 

We utilize the visual-textual data across 14 subtasks from the MME benchmark \cite{fu2023mme}, and employ JPEG compression to the images with the respective LMM-JND-guided levels. We select five representative LMMs, including Qwen2.5-VL-7B, InternVL2.5-78B, Llava-OneVision-72B, GPT-4o, and Gemini 2.0 Flash, for evaluation. Tab. \ref{JPEG} shows the response change ratio $R_{rc}$ and the average saved bits per pixel (bpp) among different perception tasks. All experiments are conducted repeatedly 3 times to avoid model hallucination.
It can be observed that injecting JPEG distortions below the $\mathcal{MRV}$ does not affect the output of GPT-4o and Gemini 2.0 Flash ($R_{rc}=0$), while effectively reducing a considerable overhead of bit rate.
For Qwen2.5-VL-7B, performing high degree of JPEG compression (causing {\it blockiness}, {\it color shifts}, and {\it ringing artifacts}, {\it etc}.) severely interferes 12 VQA evaluations with an average response change ratio of 16\%, which, together with the results obtained in Tab. \ref{exp1}, reflects the defect of Qwen-VL series in JPEG distortion perception.
Moreover, we find that LLava-OneVision-72B, despite applying a more aggressive JPEG compression scheme, demonstrates significantly higher stability in its responses compared to InternVL2.5-78B, indicating an enhanced robustness for general visual perception tasks. 
Conversely, this same characteristic implies a potential trade-off, presenting a disadvantage in scenarios that necessitate a high degree of perceptual granularity.
We also find that test sets requiring a high level of detail perception, specifically position, celebrity, and artwork, exhibit higher response change ratios than other categories.
Note that the images in the text translation set are all simple monochrome pictures with inherently small file sizes, and introducing a certain degree of JPEG compression paradoxically increases their volume.

To verify whether there exists an appropriate compression level that does not affect VQA across all models, we apply the JPEG compression (level=41) to all LMMs according to the minimum $\mathcal{MRV}$. The results show that only InternVL2.5-78B exhibits 2.7\% and 5.3\% response changes in the artwork and celebrity subsets, respectively, indicating that utilizing the common perceptual limit to achieve impact-free image compression is feasible.
In summary, LMM-JND-based visual signal compression differs from approaches guided by either conventional machine vision-JND or HVS-JND, exhibiting greater divergence among signal recipients and possessing a strong task-dependent nature.

\subsection{Beyond Visual Signals and with Other Modalities} 
In addition to pure visual perturbations, we also investigate the LMM-JND in textual inputs by quantitatively applying adversarial textual attacks. Specifically, we adapt three types of black-box attack approaches mentioned in PromptRobust \cite{zhu2023promptrobust}, including character-level, word-level, and sentence-level attacks, as follows.
\begin{itemize}
    \item {\bf Character-Level}: We introduce typos or errors to words, including adding, deleting, repeating, and permuting characters for certain words. Here, the perturbation level is defined as the proportion of manipulated characters within the entire text.
    \item {\bf Word-Level}: We modify words with synonyms or contextually similar words by a GPT-assist mode to perplex LMMs. Their proportion in the original text represents the perturbation level.
    \item {\bf Sentence-Level}: We inject irrelevant strings to the head, middle, and end of the prompts. The injected strings are composed of characters randomly sampled from ASCII codes range from 33 to 126. The ratio of their lengths to the total number of characters in the text serves as the perturbation level.
\end{itemize}

Considering the textual complexity of the attack source, we manually gather 1,000 multimodal long-context samples from 30 subjects in MMMU \cite{yue2024mmmu} with over 20 words each for the questions.
Fig. \ref{text-JND} illustrates three types of textual attack and the trend of the response change rate against different perturbation levels. First, GPT-4o is the most affected by perturbations, with its responses changing by more than 30\% in all three types of textual attacks. In comparison, Gemini 2.0 Flash achieves the lowest response variation, demonstrating good robustness and resistance to perturbations. Second, we observe several intervals where increasing the perturbation level do not significantly change the model's output variation rate, indicating the existence of LMM-JND under textual attacks.
Third, looking at the response change ratio trend over model size, we notice that larger and more instructable models become more vulnerable as they can spot the errors in prompts easily.

\section{Conclusion and Future Work}
In this paper, we demonstrate the existence of just noticeable difference (JND) in large multimodal models (LMM), termed as the LMM-JND, which statistically measures the perceptual capabilities of LMMs from the perspective of the minimal perceptible quantity.
We first define the concept of LMM-JND and then propose a LMM-oriented JND determination pipeline. Targeting various inputs that LMMs support, we construct a visual perception alignment (VPA) dataset, VPA-JND, with over 489k images to facilitate LMM-JND studies. 
Extensive experiments indicate the ubiquitous myopia or perceptual redundancy among current LMMs. 
Our analysis of how LMMs respond linguistically and visually identifies existing perception strengths ({\it e.g.}, blur, noise, and watermark) and potential risks for robustness and security concerns. Additionally, we also highlight the possible direction of improvement for LMMs in the trade-off between perception ability and security by disentangling the visual and language backbone.
Future work can be two-folds: (1) developing universal model to predict LMM-JND so as to optimize the response efficiency for LMMs; (2) LMM-JND research for cross-modality attempts such as audio, text, and image, as the ever-growing applications of LMMs.

% Acknowledgements should only appear in the accepted version.
%\section*{Acknowledgements}

%\section*{Impact Statement}

%%%%%%%%%%%%%%%%%%%%%%%%%%%%%%%%%%%%%%%%%%%%%%%%%%%%%%%%%%%%%%%%%%%%%%%%%%%%%%%
%%%%%%%%%%%%%%%%%%%%%%%%%%%%%%%%%%%%%%%%%%%%%%%%%%%%%%%%%%%%%%%%%%%%%%%%%%%%%%%
% APPENDIX
%%%%%%%%%%%%%%%%%%%%%%%%%%%%%%%%%%%%%%%%%%%%%%%%%%%%%%%%%%%%%%%%%%%%%%%%%%%%%%%
%%%%%%%%%%%%%%%%%%%%%%%%%%%%%%%%%%%%%%%%%%%%%%%%%%%%%%%%%%%%%%%%%%%%%%%%%%%%%%%
%\newpage
%\appendix
%\onecolumn

%%%%%%%%%%%%%%%%%%%%%%%%%%%%%%%%%%%%%%%%%%%%%%%%%%%%%%%%%%%%%%%%%%%%%%%%%%%%%%%
%%%%%%%%%%%%%%%%%%%%%%%%%%%%%%%%%%%%%%%%%%%%%%%%%%%%%%%%%%%%%%%%%%%%%%%%%%%%%%%


\begin{thebibliography}{10}
\providecommand{\url}[1]{#1}
\csname url@samestyle\endcsname
\providecommand{\newblock}{\relax}
\providecommand{\bibinfo}[2]{#2}
\providecommand{\BIBentrySTDinterwordspacing}{\spaceskip=0pt\relax}
\providecommand{\BIBentryALTinterwordstretchfactor}{4}
\providecommand{\BIBentryALTinterwordspacing}{\spaceskip=\fontdimen2\font plus
\BIBentryALTinterwordstretchfactor\fontdimen3\font minus \fontdimen4\font\relax}
\providecommand{\BIBforeignlanguage}[2]{{%
\expandafter\ifx\csname l@#1\endcsname\relax
\typeout{** WARNING: IEEEtran.bst: No hyphenation pattern has been}%
\typeout{** loaded for the language `#1'. Using the pattern for}%
\typeout{** the default language instead.}%
\else
\language=\csname l@#1\endcsname
\fi
#2}}
\providecommand{\BIBdecl}{\relax}
\BIBdecl

\bibitem{wang2024qwen2}
P.~Wang, S.~Bai, S.~Tan, S.~Wang, Z.~Fan, J.~Bai, K.~Chen, X.~Liu, J.~Wang, W.~Ge \emph{et~al.}, ``Qwen2-vl: Enhancing vision-language model's perception of the world at any resolution,'' \emph{arXiv preprint arXiv:2409.12191}, 2024.

\bibitem{jhamtani2018learning}
H.~Jhamtani and T.~Berg-Kirkpatrick, ``Learning to describe differences between pairs of similar images,'' in \emph{Proceedings of the 2018 Conference on Empirical Methods in Natural Language Processing}, 2018, pp. 4024--4034.

\bibitem{llama32vision}
Meta, ``Introducing llama 3.2,'' \url{https://www.llama.com}, 2024, accessed: 2024-11-30.

\bibitem{claude}
Anthropic, ``Claude opus 4,'' \url{https://www.anthropic.com/claude/opus}, 2025, accessed: 2025-06-22.

\bibitem{wu2024deepseek}
Z.~Wu, X.~Chen, Z.~Pan, X.~Liu, W.~Liu, D.~Dai, H.~Gao, Y.~Ma, C.~Wu, B.~Wang \emph{et~al.}, ``Deepseek-vl2: Mixture-of-experts vision-language models for advanced multimodal understanding,'' \emph{arXiv preprint arXiv:2412.10302}, 2024.

\bibitem{zhang2021just}
Q.~Zhang, S.~Wang, X.~Zhang, S.~Ma, and W.~Gao, ``Just recognizable distortion for machine vision oriented image and video coding,'' \emph{International Journal of Computer Vision}, vol. 129, no.~10, pp. 2889--2906, 2021.

\bibitem{zhang2018unreasonable}
R.~Zhang, P.~Isola, A.~A. Efros, E.~Shechtman, and O.~Wang, ``The unreasonable effectiveness of deep features as a perceptual metric,'' in \emph{Proceedings of the IEEE conference on computer vision and pattern recognition}, 2018, pp. 586--595.

\bibitem{jayant1993signal}
N.~Jayant, J.~Johnston, and R.~Safranek, ``Signal compression based on models of human perception,'' \emph{Proceedings of the IEEE}, vol.~81, no.~10, pp. 1385--1422, 1993.

\bibitem{zhu2023promptrobust}
K.~Zhu, J.~Wang, J.~Zhou, Z.~Wang, H.~Chen, Y.~Wang, L.~Yang, W.~Ye, Y.~Zhang, N.~Gong \emph{et~al.}, ``Promptrobust: Towards evaluating the robustness of large language models on adversarial prompts,'' in \emph{Proceedings of the 1st ACM Workshop on Large AI Systems and Models with Privacy and Safety Analysis}, 2023, pp. 57--68.

\bibitem{wang2004image}
Z.~Wang, A.~C. Bovik, H.~R. Sheikh, and E.~P. Simoncelli, ``Image quality assessment: from error visibility to structural similarity,'' \emph{IEEE transactions on image processing}, vol.~13, no.~4, pp. 600--612, 2004.

\bibitem{chen2025study}
Z.~Chen, W.~Sun, H.~Wu, Z.~Zhang, J.~Jia, R.~Huang, X.~Min, G.~Zhai, and W.~Zhang, ``Study of subjective and objective naturalness assessment of ai-generated images,'' \emph{IEEE Transactions on Circuits and Systems for Video Technology}, vol.~35, no.~4, pp. 3573--3588, 2025.

\bibitem{gemini2}
S.~Pichai, D.~Hassabis, and K.~Kavukcuoglu, ``Introducing gemini 2.0: our new ai model for the agentic era,'' \url{https://blog.google/technology/google-deepmind/google-gemini-ai-update-december-2024/#ceo-message}, 2024, accessed: 2025-3-24.

\bibitem{ansys}
Ansys, ``Ansys speos design \& validation of optical systems,'' \url{https://www.ansys.com/products/optics/ansys-speos}, 2025, accessed: 2025-1-5.

\bibitem{yang2024thinking}
J.~Yang, S.~Yang, A.~W. Gupta, R.~Han, L.~Fei-Fei, and S.~Xie, ``Thinking in space: How multimodal large language models see, remember, and recall spaces,'' \emph{arXiv preprint arXiv:2412.14171}, 2024.

\bibitem{chen2014jnd}
Z.~Chen and H.~Liu, ``Jnd modeling: Approaches and applications,'' in \emph{2014 19th International Conference on Digital Signal Processing}.\hskip 1em plus 0.5em minus 0.4em\relax IEEE, 2014, pp. 827--830.

\bibitem{qi2025shapellm}
Z.~Qi, R.~Dong, S.~Zhang, H.~Geng, C.~Han, Z.~Ge, L.~Yi, and K.~Ma, ``Shapellm: Universal 3d object understanding for embodied interaction,'' in \emph{European Conference on Computer Vision}.\hskip 1em plus 0.5em minus 0.4em\relax Springer, 2025, pp. 214--238.

\bibitem{lin2019kadid}
H.~Lin, V.~Hosu, and D.~Saupe, ``Kadid-10k: A large-scale artificially distorted iqa database,'' in \emph{2019 Eleventh International Conference on Quality of Multimedia Experience (QoMEX)}.\hskip 1em plus 0.5em minus 0.4em\relax IEEE, 2019, pp. 1--3.

\bibitem{zhang2021deep}
Y.~Zhang, H.~Liu, Y.~Yang, X.~Fan, S.~Kwong, and C.~J. Kuo, ``Deep learning based just noticeable difference and perceptual quality prediction models for compressed video,'' \emph{IEEE Transactions on Circuits and Systems for Video Technology}, vol.~32, no.~3, pp. 1197--1212, 2021.

\bibitem{liu2010just}
A.~Liu, W.~Lin, M.~Paul, C.~Deng, and F.~Zhang, ``Just noticeable difference for images with decomposition model for separating edge and textured regions,'' \emph{IEEE Transactions on Circuits and Systems for Video Technology}, vol.~20, no.~11, pp. 1648--1652, 2010.

\bibitem{liu2023medical}
F.~Liu, T.~Zhu, X.~Wu, B.~Yang, C.~You, C.~Wang, L.~Lu, Z.~Liu, Y.~Zheng, X.~Sun \emph{et~al.}, ``A medical multimodal large language model for future pandemics,'' \emph{NPJ Digital Medicine}, vol.~6, no.~1, p. 226, 2023.

\bibitem{xu2024drivegpt4}
Z.~Xu, Y.~Zhang, E.~Xie, Z.~Zhao, Y.~Guo, K.-Y.~K. Wong, Z.~Li, and H.~Zhao, ``Drivegpt4: Interpretable end-to-end autonomous driving via large language model,'' \emph{IEEE Robotics and Automation Letters}, 2024.

\bibitem{pelli2013measuring}
D.~G. Pelli and P.~Bex, ``Measuring contrast sensitivity,'' \emph{Vision research}, vol.~90, pp. 10--14, 2013.

\bibitem{brown1989high}
B.~Brown and J.~E. Lovie-Kitchin, ``High and low contrast acuity and clinical contrast sensitivity tested in a normal population,'' \emph{Optometry and vision science}, vol.~66, no.~7, pp. 467--473, 1989.

\bibitem{ahumada1992luminance}
A.~J. Ahumada~Jr and H.~A. Peterson, ``Luminance-model-based dct quantization for color image compression,'' in \emph{Human vision, visual processing, and digital display III}, vol. 1666.\hskip 1em plus 0.5em minus 0.4em\relax SPIE, 1992, pp. 365--374.

\bibitem{wei2009spatio}
Z.~Wei and K.~N. Ngan, ``Spatio-temporal just noticeable distortion profile for grey scale image/video in dct domain,'' \emph{IEEE Transactions on Circuits and Systems for Video Technology}, vol.~19, no.~3, pp. 337--346, 2009.

\bibitem{bradley1999wavelet}
A.~P. Bradley, ``A wavelet visible difference predictor,'' \emph{IEEE Transactions on image processing}, vol.~8, no.~5, pp. 717--730, 1999.

\bibitem{jiang2022toward}
Q.~Jiang, Z.~Liu, S.~Wang, F.~Shao, and W.~Lin, ``Toward top-down just noticeable difference estimation of natural images,'' \emph{IEEE Transactions on Image Processing}, vol.~31, pp. 3697--3712, 2022.

\bibitem{bae2013novel}
S.-H. Bae and M.~Kim, ``A novel dct-based jnd model for luminance adaptation effect in dct frequency,'' \emph{IEEE Signal Processing Letters}, vol.~20, no.~9, pp. 893--896, 2013.

\bibitem{goodale1992separate}
M.~A. Goodale and A.~D. Milner, ``Separate visual pathways for perception and action,'' \emph{Trends in neurosciences}, vol.~15, no.~1, pp. 20--25, 1992.

\bibitem{zhang2023magicbrush}
K.~Zhang, L.~Mo, W.~Chen, H.~Sun, and Y.~Su, ``Magicbrush: A manually annotated dataset for instruction-guided image editing,'' \emph{Advances in Neural Information Processing Systems}, vol.~36, pp. 31\,428--31\,449, 2023.

\bibitem{chen2025joint}
Z.~Chen, W.~Sun, J.~Jia, R.~Huang, F.~Lu, Y.~Chen, X.~Min, G.~Zhai, and W.~Zhang, ``Joint luminance-chrominance learning for image debanding,'' \emph{IEEE Transactions on Circuits and Systems for Video Technology}, 2025.

\bibitem{bitar2016algorithmic}
A.~W. Bitar, M.~M. Mansour, and A.~Chehab, ``Algorithmic optimizations in the hmax model targeted for efficient object recognition,'' in \emph{Computer Vision, Imaging and Computer Graphics Theory and Applications: 10th International Joint Conference, VISIGRAPP 2015, Berlin, Germany, March 11--14, 2015, Revised Selected Papers 10}.\hskip 1em plus 0.5em minus 0.4em\relax Springer, 2016, pp. 374--395.

\bibitem{ficsek2023cortico}
M.~Fi{\c{s}}ek, D.~Herrmann, A.~Egea-Weiss, M.~Cloves, L.~Bauer, T.-Y. Lee, L.~E. Russell, and M.~H{\"a}usser, ``Cortico-cortical feedback engages active dendrites in visual cortex,'' \emph{Nature}, vol. 617, no. 7962, pp. 769--776, 2023.

\bibitem{grill2004human}
K.~Grill-Spector and R.~Malach, ``The human visual cortex,'' \emph{Annu. Rev. Neurosci.}, vol.~27, no.~1, pp. 649--677, 2004.

\bibitem{yue2024mmmu}
X.~Yue, Y.~Ni, K.~Zhang, T.~Zheng, R.~Liu, G.~Zhang, S.~Stevens, D.~Jiang, W.~Ren, Y.~Sun \emph{et~al.}, ``Mmmu: A massive multi-discipline multimodal understanding and reasoning benchmark for expert agi,'' in \emph{Proceedings of the IEEE/CVF Conference on Computer Vision and Pattern Recognition}, 2024, pp. 9556--9567.

\bibitem{milner2017two}
A.~D. Milner, ``How do the two visual streams interact with each other?'' \emph{Experimental brain research}, vol. 235, pp. 1297--1308, 2017.

\bibitem{chen2024obi}
Z.~Chen, T.~Chen, W.~Zhang, and G.~Zhai, ``Obi-bench: Can lmms aid in study of ancient script on oracle bones?'' \emph{arXiv preprint arXiv:2412.01175}, 2024.

\bibitem{mao2023transfer}
Y.~Mao, J.~Wu, X.~Wang, L.~Li, and W.~Dong, ``Transfer learning for just noticeable difference estimation,'' \emph{Information Sciences}, vol. 648, p. 119575, 2023.

\bibitem{yuan2019visual}
D.~Yuan, T.~Zhao, Y.~Xu, H.~Xue, and L.~Lin, ``Visual jnd: A perceptual measurement in video coding,'' \emph{IEEE Access}, vol.~7, pp. 29\,014--29\,022, 2019.

\bibitem{chou1995perceptually}
C.-H. Chou and Y.-C. Li, ``A perceptually tuned subband image coder based on the measure of just-noticeable-distortion profile,'' \emph{IEEE Transactions on circuits and systems for video technology}, vol.~5, no.~6, pp. 467--476, 1995.

\bibitem{zhang2024perceptual}
Q.~Zhang, S.~Wang, X.~Zhang, C.~Jia, Z.~Wang, S.~Ma, and W.~Gao, ``Perceptual video coding for machines via satisfied machine ratio modeling,'' \emph{IEEE Transactions on Pattern Analysis and Machine Intelligence}, vol.~46, no.~12, pp. 7651--7668, 2024.

\bibitem{jin2022just}
J.~Jin, X.~Zhang, X.~Fu, H.~Zhang, W.~Lin, J.~Lou, and Y.~Zhao, ``Just noticeable difference for deep machine vision,'' \emph{IEEE Transactions on Circuits and Systems for Video Technology}, vol.~32, no.~6, pp. 3452--3461, 2022.

\bibitem{tu2020bband}
Z.~Tu, J.~Lin, Y.~Wang, B.~Adsumilli, and A.~C. Bovik, ``Bband index: A no-reference banding artifact predictor,'' in \emph{ICASSP 2020-2020 IEEE International Conference on Acoustics, Speech and Signal Processing (ICASSP)}.\hskip 1em plus 0.5em minus 0.4em\relax IEEE, 2020, pp. 2712--2716.

\bibitem{bt2002methodology}
R.~BT, ``Methodology for the subjective assessment of the quality of television pictures,'' \emph{International Telecommunication Union}, vol.~4, p.~19, 2002.

\bibitem{williams2018broad}
A.~Williams, N.~Nangia, and S.~Bowman, ``A broad-coverage challenge corpus for sentence understanding through inference,'' in \emph{Proceedings of the 2018 Conference of the North American Chapter of the Association for Computational Linguistics: Human Language Technologies, Volume 1 (Long Papers)}, 2018, pp. 1112--1122.

\bibitem{lewis2020bart}
M.~Lewis, Y.~Liu, N.~Goyal, M.~Ghazvininejad, A.~Mohamed, O.~Levy, V.~Stoyanov, and L.~Zettlemoyer, ``Bart: Denoising sequence-to-sequence pre-training for natural language generation, translation, and comprehension,'' in \emph{Proceedings of the 58th Annual Meeting of the Association for Computational Linguistics}, 2020, pp. 7871--7880.

\bibitem{chen2024fs}
Z.~Chen, W.~Sun, Z.~Zhang, R.~Huang, F.~Lu, X.~Min, G.~Zhai, and W.~Zhang, ``Fs-band: A frequency-sensitive banding detector,'' in \emph{2024 IEEE International Symposium on Circuits and Systems (ISCAS)}.\hskip 1em plus 0.5em minus 0.4em\relax IEEE, 2024, pp. 1--5.

\bibitem{chen2024band}
Z.~Chen, W.~Sun, J.~Jia, F.~Lu, Z.~Zhang, J.~Liu, R.~Huang, X.~Min, and G.~Zhai, ``Band-2k: Banding artifact noticeable database for banding detection and quality assessment,'' \emph{IEEE Transactions on Circuits and Systems for Video Technology}, vol.~34, no.~7, pp. 6347--6362, 2024.

\bibitem{wang2017videoset}
H.~Wang, I.~Katsavounidis, J.~Zhou, J.~Park, S.~Lei, X.~Zhou, M.-O. Pun, X.~Jin, R.~Wang, X.~Wang \emph{et~al.}, ``Videoset: A large-scale compressed video quality dataset based on jnd measurement,'' \emph{Journal of Visual Communication and Image Representation}, vol.~46, pp. 292--302, 2017.

\bibitem{jin2016statistical}
L.~Jin, J.~Y. Lin, S.~Hu, H.~Wang, P.~Wang, I.~Katsavounidis, A.~Aaron, and C.-C.~J. Kuo, ``Statistical study on perceived jpeg image quality via mcl-jci dataset construction and analysis,'' \emph{Electronic Imaging}, vol. 2016, no.~13, pp. 1--9, 2016.

\bibitem{shen2020just}
X.~Shen, Z.~Ni, W.~Yang, X.~Zhang, S.~Wang, and S.~Kwong, ``Just noticeable distortion profile inference: A patch-level structural visibility learning approach,'' \emph{IEEE Transactions on Image Processing}, vol.~30, pp. 26--38, 2020.

\bibitem{sheng2024audio}
N.~Sheng, H.~Yin, H.~Wang, L.~Mo, Y.~Liu, X.~Huang, J.~Lin, and X.~Tang, ``Audio--video collaborative jnd estimation model for multimedia applications,'' \emph{Journal of Visual Communication and Image Representation}, vol. 103, p. 104254, 2024.

\bibitem{weber1996eh}
E.~H. Weber, \emph{EH Weber on the tactile senses}.\hskip 1em plus 0.5em minus 0.4em\relax Psychology Press, 1996.

\bibitem{cui2024robustness}
X.~Cui, A.~Aparcedo, Y.~K. Jang, and S.-N. Lim, ``On the robustness of large multimodal models against image adversarial attacks,'' in \emph{Proceedings of the IEEE/CVF Conference on Computer Vision and Pattern Recognition}, 2024, pp. 24\,625--24\,634.

\bibitem{tong2024eyes}
S.~Tong, Z.~Liu, Y.~Zhai, Y.~Ma, Y.~LeCun, and S.~Xie, ``Eyes wide shut? exploring the visual shortcomings of multimodal llms,'' in \emph{Proceedings of the IEEE/CVF Conference on Computer Vision and Pattern Recognition}, 2024, pp. 9568--9578.

\bibitem{yang2025visionzip}
S.~Yang, Y.~Chen, Z.~Tian, C.~Wang, J.~Li, B.~Yu, and J.~Jia, ``Visionzip: Longer is better but not necessary in vision language models,'' in \emph{Proceedings of the Computer Vision and Pattern Recognition Conference}, 2025, pp. 19\,792--19\,802.

\bibitem{zhang2023learning}
Y.~Zhang, H.~Lin, J.~Sun, L.~Zhu, and S.~Kwong, ``Learning to predict object-wise just recognizable distortion for image and video compression,'' \emph{IEEE Transactions on Multimedia}, vol.~26, pp. 5925--5938, 2023.

\bibitem{liu2019deep}
H.~Liu, Y.~Zhang, H.~Zhang, C.~Fan, S.~Kwong, C.-C.~J. Kuo, and X.~Fan, ``Deep learning-based picture-wise just noticeable distortion prediction model for image compression,'' \emph{IEEE Transactions on Image Processing}, vol.~29, pp. 641--656, 2019.

\bibitem{lin2022large}
H.~Lin, G.~Chen, M.~Jenadeleh, V.~Hosu, U.-D. Reips, R.~Hamzaoui, and D.~Saupe, ``Large-scale crowdsourced subjective assessment of picturewise just noticeable difference,'' \emph{IEEE transactions on circuits and systems for video technology}, vol.~32, no.~9, pp. 5859--5873, 2022.

\bibitem{zhu2025internvl3}
J.~Zhu, W.~Wang, Z.~Chen, Z.~Liu, S.~Ye, L.~Gu, H.~Tian, Y.~Duan, W.~Su, J.~Shao \emph{et~al.}, ``Internvl3: Exploring advanced training and test-time recipes for open-source multimodal models,'' \emph{arXiv preprint arXiv:2504.10479}, 2025.

\bibitem{carlson2016clinical}
N.~B. Carlson, D.~Kurtz, and C.~Hines, \emph{Clinical procedures for ocular examination}.\hskip 1em plus 0.5em minus 0.4em\relax McGraw-Hill Education, 2016.

\bibitem{chen2024gaia}
Z.~Chen, W.~Sun, Y.~Tian, J.~Jia, Z.~Zhang, W.~Jiarui, R.~Huang, X.~Min, G.~Zhai, and W.~Zhang, ``Gaia: Rethinking action quality assessment for ai-generated videos,'' in \emph{Advances in Neural Information Processing Systems}, vol.~37, 2024, pp. 40\,111--40\,144.

\bibitem{chen2024expanding}
Z.~Chen, W.~Wang, Y.~Cao, Y.~Liu, Z.~Gao, E.~Cui, J.~Zhu, S.~Ye, H.~Tian, Z.~Liu \emph{et~al.}, ``Expanding performance boundaries of open-source multimodal models with model, data, and test-time scaling,'' \emph{arXiv preprint arXiv:2412.05271}, 2024.

\bibitem{shen2003foundations}
J.~Shen, ``On the foundations of vision modeling: I. weber’s law and weberized tv restoration,'' \emph{Physica D: Nonlinear Phenomena}, vol. 175, no. 3-4, pp. 241--251, 2003.

\bibitem{fu2023mme}
C.~Fu, P.~Chen, Y.~Shen, Y.~Qin, M.~Zhang, X.~Lin, J.~Yang, X.~Zheng, K.~Li, X.~Sun \emph{et~al.}, ``Mme: A comprehensive evaluation benchmark for multimodal large language models,'' \emph{arXiv preprint arXiv:2306.13394}, 2023.

\bibitem{bai2025qwen25vl}
\BIBentryALTinterwordspacing
S.~Bai, K.~Chen, X.~Liu, J.~Wang, W.~Ge, S.~Song, K.~Dang, P.~Wang, S.~Wang, J.~Tang, H.~Zhong, Y.~Zhu, M.~Yang, Z.~Li, J.~Wan, P.~Wang, W.~Ding, Z.~Fu, Y.~Xu, J.~Ye, X.~Zhang, T.~Xie, Z.~Cheng, H.~Zhang, Z.~Yang, H.~Xu, and J.~Lin, ``Qwen2.5-vl technical report,'' 2025. [Online]. Available: \url{https://arxiv.org/abs/2502.13923}
\BIBentrySTDinterwordspacing

\bibitem{SmolVLM2025}
A.~Marafioti, M.~Noyan, M.~Farré, E.~Bakouch, and P.~Cuenca, ``Smolvlm - small yet mighty vision language model,'' \url{https://huggingface.co/blog/smolvlm}, 2024, accessed: 2025-02-23.

\bibitem{liu2023first}
Y.~Liu, J.~Jin, Y.~Xue, and W.~Lin, ``The first comprehensive dataset with multiple distortion types for visual just-noticeable differences,'' in \emph{2023 IEEE International Conference on Image Processing (ICIP)}.\hskip 1em plus 0.5em minus 0.4em\relax IEEE, 2023, pp. 2820--2824.

\bibitem{fan2019picture}
C.~Fan, Y.~Zhang, H.~Zhang, R.~Hamzaoui, and Q.~Jiang, ``Picture-level just noticeable difference for symmetrically and asymmetrically compressed stereoscopic images: Subjective quality assessment study and datasets,'' \emph{Journal of Visual Communication and Image Representation}, vol.~62, pp. 140--151, 2019.

\bibitem{huang2017measure}
Q.~Huang, H.~Wang, S.~C. Lim, H.~Y. Kim, S.~Y. Jeong, and C.-C.~J. Kuo, ``Measure and prediction of hevc perceptually lossy/lossless boundary qp values,'' in \emph{2017 data compression conference (DCC)}.\hskip 1em plus 0.5em minus 0.4em\relax IEEE, 2017, pp. 42--51.

\bibitem{yin2024survey}
\BIBentryALTinterwordspacing
S.~Yin, C.~Fu, S.~Zhao, K.~Li, X.~Sun, T.~Xu, and E.~Chen, ``A survey on multimodal large language models,'' \emph{National Science Review}, vol.~11, no.~12, p. nwae403, 11 2024. [Online]. Available: \url{https://doi.org/10.1093/nsr/nwae403}
\BIBentrySTDinterwordspacing

\bibitem{lin2021progress}
W.~Lin and G.~Ghinea, ``Progress and opportunities in modelling just-noticeable difference (jnd) for multimedia,'' \emph{IEEE Transactions on Multimedia}, vol.~24, pp. 3706--3721, 2021.

\bibitem{testolina2023jpeg}
M.~Testolina, V.~Hosu, M.~Jenadeleh, D.~Lazzarotto, D.~Saupe, and T.~Ebrahimi, ``Jpeg aic-3 dataset: towards defining the high quality to nearly visually lossless quality range,'' in \emph{2023 15th International Conference on Quality of Multimedia Experience (QoMEX)}.\hskip 1em plus 0.5em minus 0.4em\relax IEEE, 2023, pp. 55--60.

\bibitem{klein2023r3vival}
F.~Klein and S.~V.~A. Gar{\'\i}, ``The r3vival dataset: Repository of room responses and 360 videos of a variable acoustics lab,'' in \emph{ICASSP 2023-2023 IEEE International Conference on Acoustics, Speech and Signal Processing (ICASSP)}.\hskip 1em plus 0.5em minus 0.4em\relax IEEE, 2023, pp. 1--5.

\bibitem{liu2018jnd}
X.~Liu, Z.~Chen, X.~Wang, J.~Jiang, and S.~Kowng, ``Jnd-pano: Database for just noticeable difference of jpeg compressed panoramic images,'' in \emph{Advances in Multimedia Information Processing--PCM 2018: 19th Pacific-Rim Conference on Multimedia, Hefei, China, September 21-22, 2018, Proceedings, Part I 19}.\hskip 1em plus 0.5em minus 0.4em\relax Springer, 2018, pp. 458--468.

\bibitem{wu2013perceptual}
H.~R. Wu, A.~R. Reibman, W.~Lin, F.~Pereira, and S.~S. Hemami, ``Perceptual visual signal compression and transmission,'' \emph{Proceedings of the IEEE}, vol. 101, no.~9, pp. 2025--2043, 2013.

\bibitem{wu2019survey}
J.~Wu, G.~Shi, and W.~Lin, ``Survey of visual just noticeable difference estimation,'' \emph{Frontiers of Computer Science}, vol.~13, pp. 4--15, 2019.

\bibitem{bai2023qwen}
J.~Bai, S.~Bai, S.~Yang, S.~Wang, S.~Tan, P.~Wang, J.~Lin, C.~Zhou, and J.~Zhou, ``Qwen-vl: A versatile vision-language model for understanding, localization, text reading, and beyond,'' \emph{arXiv preprint arXiv:2308.12966}, vol.~1, no.~2, p.~3, 2023.

\bibitem{ak2022just}
A.~Ak, A.~Pastor, and P.~Le~Callet, ``From just noticeable differences to image quality,'' in \emph{Proceedings of the 2nd Workshop on Quality of Experience in Visual Multimedia Applications}, 2022, pp. 23--28.

\bibitem{wu2024qbench}
H.~Wu, Z.~Zhang, E.~Zhang, C.~Chen, L.~Liao, A.~Wang, C.~Li, W.~Sun, Q.~Yan, G.~Zhai \emph{et~al.}, ``Q-bench: A benchmark for general-purpose foundation models on low-level vision,'' in \emph{The Twelfth International Conference on Learning Representations}, 2024.

\bibitem{zhang2024mm}
D.~Zhang, Y.~Yu, J.~Dong, C.~Li, D.~Su, C.~Chu, and D.~Yu, ``Mm-llms: Recent advances in multimodal large language models,'' \emph{arXiv preprint arXiv:2401.13601}, 2024.

\bibitem{wu2023multimodal}
J.~Wu, W.~Gan, Z.~Chen, S.~Wan, and S.~Y. Philip, ``Multimodal large language models: A survey,'' in \emph{2023 IEEE International Conference on Big Data (BigData)}.\hskip 1em plus 0.5em minus 0.4em\relax IEEE, 2023, pp. 2247--2256.

\bibitem{wang2016mcl}
H.~Wang, W.~Gan, S.~Hu, J.~Y. Lin, L.~Jin, L.~Song, P.~Wang, I.~Katsavounidis, A.~Aaron, and C.-C.~J. Kuo, ``Mcl-jcv: a jnd-based h. 264/avc video quality assessment dataset,'' in \emph{2016 IEEE international conference on image processing (ICIP)}.\hskip 1em plus 0.5em minus 0.4em\relax IEEE, 2016, pp. 1509--1513.

\bibitem{wan2020pattern}
W.~Wan, J.~Wang, J.~Li, L.~Meng, J.~Sun, H.~Zhang, and J.~Liu, ``Pattern complexity-based jnd estimation for quantization watermarking,'' \emph{Pattern Recognition Letters}, vol. 130, pp. 157--164, 2020.

\bibitem{flynn2013image}
J.~R. Flynn, S.~Ward, J.~Abich, and D.~Poole, ``Image quality assessment using the ssim and the just noticeable difference paradigm,'' in \emph{Engineering Psychology and Cognitive Ergonomics. Understanding Human Cognition: 10th International Conference, EPCE 2013, Held as Part of HCI International 2013, Las Vegas, NV, USA, July 21-26, 2013, Proceedings, Part I 10}.\hskip 1em plus 0.5em minus 0.4em\relax Springer, 2013, pp. 23--30.

\bibitem{wang2020hierarchical}
H.~Wang, L.~Yu, J.~Liang, H.~Yin, T.~Li, and S.~Wang, ``Hierarchical predictive coding-based jnd estimation for image compression,'' \emph{IEEE Transactions on Image Processing}, vol.~30, pp. 487--500, 2020.

\bibitem{cao2024sg}
L.~Cao, W.~Sun, X.~Min, J.~Jia, Z.~Zhang, Z.~Chen, Y.~Zhu, L.~Liu, Q.~Chen, J.~Chen \emph{et~al.}, ``Sg-jnd: Semantic-guided just noticeable distortion predictor for image compression,'' in \emph{2024 IEEE International Conference on Image Processing (ICIP)}.\hskip 1em plus 0.5em minus 0.4em\relax IEEE, 2024, pp. 1139--1145.

\bibitem{reid2024gemini}
M.~Reid, N.~Savinov, D.~Teplyashin, D.~Lepikhin, T.~Lillicrap, J.-b. Alayrac, R.~Soricut, A.~Lazaridou, O.~Firat, J.~Schrittwieser \emph{et~al.}, ``Gemini 1.5: Unlocking multimodal understanding across millions of tokens of context,'' \emph{arXiv preprint arXiv:2403.05530}, 2024.

\bibitem{GPT-4o}
OpenAI, ``Hello gpt-4o,'' \url{https://openai.com/index/hello-gpt-4o/}, 2024, accessed: 2024-08-27.

\bibitem{gemini2.5}
S.~B. Mallick and L.~Kilpatrick, ``Gemini 2.5: Updates to our family of thinking models,'' \url{https://developers.googleblog.com/en/gemini-2-5-thinking-model-updates/}, 2025, accessed: 2025-06-21.

\bibitem{li2024llava}
B.~Li, Y.~Zhang, D.~Guo, R.~Zhang, F.~Li, H.~Zhang, K.~Zhang, P.~Zhang, Y.~Li, Z.~Liu \emph{et~al.}, ``Llava-onevision: Easy visual task transfer,'' \emph{arXiv preprint arXiv:2408.03326}, 2024.

\end{thebibliography}
\end{document}